%% file: neurips_2026.tex
\documentclass{article}

\PassOptionsToPackage{numbers, compress}{natbib}

\usepackage[main, final]{neurips_2026}
\usepackage[utf8]{inputenc} 
\usepackage[T1]{fontenc}    

\usepackage{graphicx}
\usepackage{subcaption}
\usepackage{amsmath}
\usepackage{multirow}
\usepackage{wrapfig}
\usepackage{hyperref}       
\usepackage{url}            
\usepackage{booktabs}       
\usepackage{amsfonts}       
\usepackage{nicefrac}       
\usepackage{microtype}      
\usepackage{xcolor}         
\usepackage{enumitem}

\title{When LLM Routers Overpay: The Cost of Strong-Model Preference}

\author{%
  Guannan Lai\textsuperscript{1,2}
  \quad
  Long Chen\textsuperscript{3}
  \quad
  Han-Jia Ye\textsuperscript{1,2}\thanks{Corresponding author.}
  \\[1.5mm]
  \textsuperscript{1}School of Artificial Intelligence, Nanjing University \\
  \textsuperscript{2}National Key Laboratory for Novel Software Technology, Nanjing University \\
  \textsuperscript{3}The Hong Kong University of Science and Technology \\
  \texttt{\{laign, yehj\}@lamda.nju.edu.cn}%
  \quad \texttt{longchen@ust.hk}
}

\begin{document}

\maketitle

\begin{abstract}
\input{tex/0_abs}
\end{abstract}

\input{tex/1_intro}
\input{tex/2_rw}
\input{tex/3_ana}
\input{tex/4_method}
\input{tex/5_exper}

\input{tex/6_con}

\begin{ack}
This work was supported by the National Key R\&D Program of China (Grant No. 2024YFE0202800) and the National Natural Science Foundation of China (Grant Nos. 62522605 and 62376118). Long Chen was additionally supported by the National Natural Science Foundation of China (Grant Nos. 62522216 and 62402408) and the Research Grants Council of the Hong Kong SAR (Grant Nos. 26208924 and 16219025).
\end{ack}

\section*{Competing Interests Disclosure}
The authors declare that they have no competing interests
that could have influenced the research, results, or
conclusions presented in this paper.

\bibliographystyle{plainnat}
\bibliography{references}


\clearpage
\appendix

\section*{Appendix}
\input{tex/7_appendix}


\newpage
\input{checklist.tex}

\end{document}

%% file: tex/0_abs.tex
LLM routing aims to reduce inference costs by assigning easy queries to less expensive models and reserving stronger models for cases in which they are needed. However, we find that existing routers often violate this principle under loose cost budgets: as the budget increases, they increasingly select the strongest and most expensive model even when less expensive feasible candidates provide comparable or identical utility. We refer to this behavior as \emph{strong-model over-selection under loose budgets}. Our analysis shows that ambiguity alone does not explain this phenomenon. Substantial overpayment can arise when differently priced models offer equivalent or near-equivalent utility and the router retains a persistent model-level preference for costly models; budget relaxation exposes this preference by making premium candidates feasible. Motivated by this diagnosis, we propose \textbf{EquiRouter}, a decision-aligned router that learns query-dependent model rankings, preserves meaningful utility differences, and explicitly prefers lower-cost models among equivalent candidates through a Cost-Aware Ranking Objective and a lightweight Query--Model Interaction Representation. Experiments across diverse text, multimodal, continuous-utility, and large-model-pool settings show that EquiRouter consistently improves cost efficiency while maintaining competitive peak and whole-curve performance. Code is available at \url{https://github.com/LAMDA-Model-Reuse/EquiRouter}.

%% file: tex/1_intro.tex
\section{Introduction}

\begin{figure*}[t]
    \centering
    \begin{subfigure}[t]{0.48\linewidth}
        \centering
        \includegraphics[width=\linewidth]{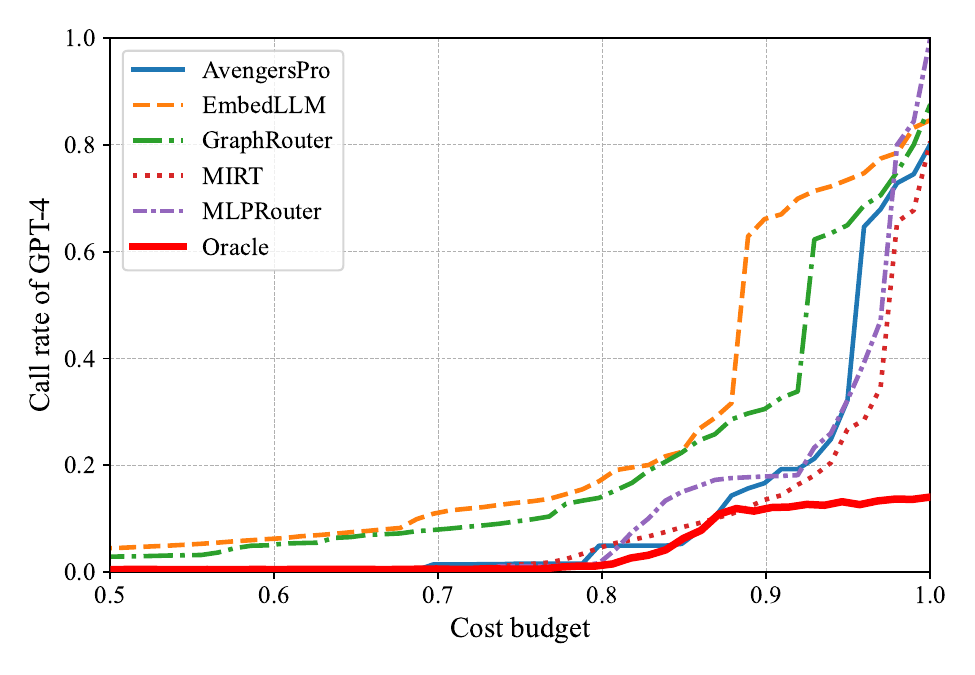}
        \caption{Strongest-model call rate under increasing budgets.}
        \label{subfig:intro_smo}
    \end{subfigure}
    \hfill
    \begin{subfigure}[t]{0.48\linewidth}
        \centering
        \includegraphics[width=\linewidth]{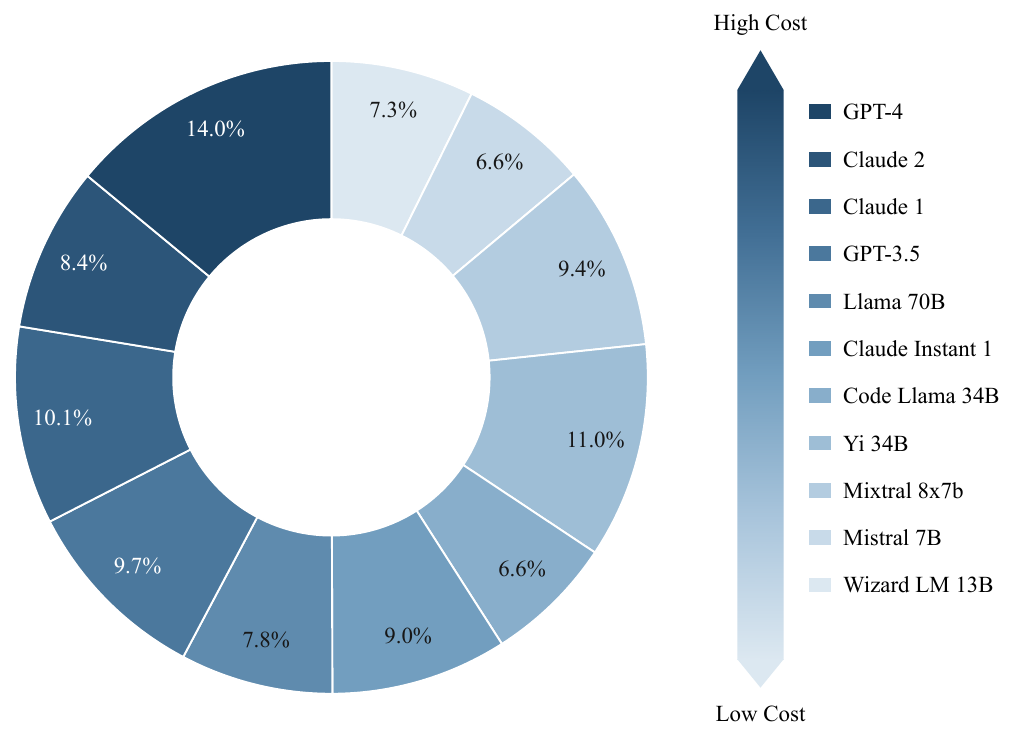}
        \caption{Oracle per-model call frequency on RouterBench.}
        \label{subfig:intro_oracle}
    \end{subfigure}
    \caption{
    Illustration of \textbf{strong-model over-selection under loose budgets} on RouterBench~\citep{hu2024routerbench}.
    \textbf{Left:} As the cost budget increases, existing routers rapidly increase their call rate to the strongest model, GPT-4, and several methods nearly saturate at 100\%.
    \textbf{Right:} In contrast, the Oracle, which has access to per-query outcomes and applies cost-based tie breaking, selects GPT-4 for fewer than 20\% of queries and continues to use cheaper models for a substantial fraction of cases.
    }
    \label{fig:intro}
    \vspace{-0.5cm}
\end{figure*}

Large language models (LLMs) deliver state-of-the-art performance across many NLP tasks, but their substantial inference cost hinders scalable deployment
\cite{bang2023gptcache,chen2024role,chen2026peakbench,lai2026pretrain,li2026consistent}.
A widely adopted mitigation is \emph{LLM routing}, where a lightweight router assigns each incoming query to an appropriate model from a heterogeneous pool, sending easy queries to small, inexpensive models and hard queries to larger, more capable ones
\cite{aggarwal2024automix,zhao2024eagle,lai2026routing}.
The goal of routing is therefore not merely to achieve high task performance, but also to avoid unnecessary calls to expensive models whenever cheaper candidates can provide comparable utility.
Most existing approaches operationalize this goal by learning query-dependent estimates of model performance and then using them to make cost-aware routing decisions
\cite{ding2025best,feng2025graphrouter,lai2026sampled}.

However, we find that existing routers often exhibit a cost-inefficient behavior under loose budgets: as the user budget increases, they increasingly select the most powerful and expensive model even when cheaper models already achieve comparable outcomes.
As shown in Fig.~\ref{fig:intro}, RouterBench contains 11 models with diverse capabilities and costs, yet several existing routers route almost all queries to GPT-4 as the budget grows.
In contrast, the Oracle continues to use cheaper models for a substantial fraction of queries.
This gap indicates that current routers may underutilize cheaper models and incur avoidable computation and monetary costs.
We refer to this behavior as \emph{strong-model over-selection under loose budgets}: routers increasingly rely on the strongest model not only when it is uniquely necessary, but also when cheaper feasible models can provide comparable utility.
We observe this pattern across multiple routing settings, making it an important obstacle to effective cost control.
We further investigate whether this phenomenon extends beyond
conventionally trained routers.
Our analysis of a System One decision model adapted to LLM routing
reveals similar cost inefficiencies, suggesting that utility-oriented
decision-making alone does not necessarily prevent overpayment.

To understand why this behavior emerges, we first examine whether it can be attributed to poor \textbf{generalization}.
Even when routers are trained and evaluated on the same queries, the qualitative over-selection pattern persists across a wide range of budgets.
This indicates that ordinary train--test distribution shift alone is insufficient to explain the observed behavior and motivates a closer examination of how routers resolve competing feasible choices.

Our analysis reveals two complementary ingredients.
First, many routing instances contain differently priced models with equivalent or near-equivalent utility, creating opportunities for quality-preserving overpayment.
Second, trained routers can retain a persistent model-level preference toward premium models within such ambiguous choices.
Controlled experiments show that symmetric uncertainty alone produces decision instability but no systematic premium direction, whereas a small premium-aligned model component consistently shifts selections toward costly models.
As the budget increases, more premium candidates become feasible, allowing this directional preference to translate into realized overpayment.

Motivated by this diagnosis, we propose \textbf{EquiRouter}, a decision-aligned routing framework that directly learns query-dependent model rankings.
Rather than suppressing strong models globally, EquiRouter preserves utility-driven preferences when model quality differs meaningfully and introduces an explicit cost preference only among equivalent or near-equivalent candidates.
It combines a Cost-Aware Ranking Objective with a lightweight Query--Model Interaction Representation, enabling query-specific model comparisons while preserving parameter sharing and inference efficiency.

We evaluate EquiRouter on RouterBench~\cite{hu2024routerbench},
MMR-Bench~\cite{ma2026mmrbench},
MixInstruct~\cite{jiang2023llm}, and
RouterEval~\cite{huang2025routereval}.
Across text-only, multimodal, continuous-utility, and large-model-pool routing settings, EquiRouter consistently improves target-level cost efficiency while maintaining competitive peak and whole-curve performance.
Ablation studies further isolate the contributions of query--model interaction and cost-aware tie-breaking.
We additionally benchmark two recent System One decision models,
Jev 1.14 and Clef-Flash, under the same budget-constrained
routing protocol, enabling a unified quality--cost comparison
with conventional learned routers.

The main contributions of this work are summarized as follows:

\begin{itemize}
[leftmargin=*]
\item We identify and characterize \emph{strong-model over-selection under loose budgets}, a cost-inefficient behavior in which LLM routers increasingly select expensive models even when cheaper feasible candidates provide comparable utility.

\item We provide an empirical diagnosis showing that substantial overpayment arises from the interaction between quality-equivalent routing opportunities and persistent cost-aligned model preferences; symmetric uncertainty alone explains decision sensitivity but not its direction.

\item We introduce \textbf{EquiRouter}, a decision-aligned routing framework that preserves meaningful utility ordering while introducing cost-aware preferences among equivalent candidates, leading to improved quality--cost trade-offs across diverse routing settings.

\end{itemize}

%% file: tex/2_rw.tex
\section{Related Work}

Multi-LLM routing has recently attracted growing attention as a general framework for reducing inference cost while maintaining high task performance \citep{srivatsa2024harnessing,varangot2025doing}.
Most existing approaches train a lightweight router to estimate model performance or routing preferences from the input query, and then select a model under a user-specified budget \citep{ong2024routellm,stripelis2024tensoropera}.
Our work follows this cost-aware routing setting, but focuses on a behavior often overlooked by aggregate quality--cost metrics: under loose budgets, routers may systematically favor expensive models even when cheaper feasible alternatives provide comparable utility.

\noindent \textbf{Heuristic- and prediction-based routing} \quad
Early LLM routing methods often focus on two-model settings and rely on hand-crafted signals, such as confidence estimates, uncertainty scores, or benchmark-derived rules, to decide whether a query should be handled by a weaker or stronger model \citep{zhang2025leveraging,shnitzer2023llm}.
Although simple and lightweight, such heuristics are difficult to scale to heterogeneous model pools with diverse capabilities and costs.

A larger body of work formulates routing as supervised prediction.
In two-model settings, routers commonly estimate the expected performance or utility of candidate models and then make a cost-aware selection \citep{ding2024hybrid,ong2024routellm}.
This paradigm has been extended to multi-model routing with kNN- or MLP-based predictors, as in \textit{TensorOpera} and \textit{Tryage} \citep{hari2023tryage,stripelis2024tensoropera}, linear regression-based routers such as \textit{MixLLM} \citep{wang2025mixllm}, clustering-based methods built on correctness-oriented representations \citep{jitkrittum2025universal,zhang2025beyond}, and structured predictors such as \textit{GraphRouter} \citep{feng2025graphrouter}.
Other methods align model behaviors in a shared representation space, as in \textit{RouterDC} \citep{chen2024routerdc}, use LLMs themselves as routers \citep{mohammadshahi2024routoo,wang2025icl,zhang2025capability}, or introduce causal formulations for model selection \citep{tsiourvas2025causal}.
While these approaches achieve practical quality--cost gains, our analysis reveals that learned routing scores can retain persistent model-level preferences toward costly models.
Such preferences may be useful for capturing global capability differences, but can induce systematic overpayment when cheaper models provide equivalent utility and no explicit cost preference resolves the choice.

\noindent \textbf{Ranking-based routing and model selection} \quad
Recent work has explored ranking or preference signals for LLM selection.
\textit{RouteLLM} learns preference-based routing decisions between candidate models \citep{ong2024routellm}, \textit{LLMRank} combines cost-penalized utility targets with pointwise and listwise ranking supervision \citep{agrawal2025llmrank}, and \textit{LLM-Blender} ranks or combines candidate model outputs to improve response quality \citep{jiang2023llm}.
These methods demonstrate the value of relative comparisons for LLM selection.
EquiRouter differs in how cost enters the ranking objective.
Unlike cost-penalized objectives that combine utility and cost through scalarization, EquiRouter follows a utility-first preference rule: meaningful utility differences determine the ordering, while cost explicitly resolves equivalent or near-equivalent choices.
This targeted supervision preserves useful capability information without systematically favoring expensive models when cheaper alternatives are sufficient.

\noindent \textbf{Routing benchmarks} \quad
Several benchmarks support systematic evaluation of LLM routing.
RouterBench \citep{hu2024routerbench} provides a widely used text-only benchmark with cached model outcomes and costs, while RouterEval \citep{huang2025routereval} offers a larger-scale evaluation suite with many candidate models.
MMR-Bench \citep{ma2026mmrbench} extends routing evaluation to multimodal settings involving vision-language tasks.
We evaluate on these benchmarks together with MixInstruct \citep{jiang2023llm} to cover text-only routing, multimodal routing, continuous-utility evaluation, and larger model pools.

%% file: tex/3_ana.tex
\section{Understanding Strong-Model Over-Selection}
\label{sec:3}

We next analyze why learned routers may over-select expensive models as the
budget increases. Our analysis separates two factors: whether the candidate
set contains cheaper models with comparable utility, and whether the router
systematically prefers the more expensive model in such cases.

\subsection{Problem Definition}
\label{sec:analysis_definition}

For a query $x$ and budget $B$, let
\[
\mathcal{F}(x,B)=\{i:c_i(x)\le B\}
\]
denote the feasible model set, where $u_i(x)$ and $c_i(x)$ are the observed
utility and inference cost of model $i$. Let
\[
u^*(x,B)=\max_{i\in\mathcal{F}(x,B)}u_i(x).
\]
We define the set of acceptable models as
\begin{equation}
\mathcal{A}_{\delta}(x,B)
=
\{i\in\mathcal{F}(x,B):
u_i(x)\ge u^*(x,B)-\delta\},
\label{eq:acceptable} 
\end{equation}
where $\delta=0$ for discrete correctness metrics and a small
validation-selected tolerance is used for continuous utilities.

For a router decision $R(x,B)$, we say that it incurs
\emph{quality-preserving overpayment} if it selects an acceptable model that
is more expensive than the cheapest acceptable alternative:
\begin{equation}
O_{\delta}(x,B)
=
\mathbf{1}\!\left[
R(x,B)\in\mathcal{A}_{\delta}(x,B),\
c_{R(x,B)}>
\min_{i\in\mathcal{A}_{\delta}(x,B)}c_i
\right].
\label{eq:overpay}
\end{equation}
This definition distinguishes avoidable cost from a quality error, i.e.,
$R(x,B)\notin\mathcal{A}_{\delta}(x,B)$.

Let $H_{\delta}(x,B)$ indicate that
$\mathcal{A}_{\delta}(x,B)$ contains models with different costs.
Since overpayment is possible only when such an opportunity exists,
\begin{equation}
\Pr(O_{\delta}=1)
=
\Pr(H_{\delta}=1)
\Pr(O_{\delta}=1\mid H_{\delta}=1).
\label{eq:decomposition}
\end{equation}
The first term is determined by the benchmark and the budget, while the
second measures the router's preference among quality-equivalent choices.

To characterize this preference, decompose any comparable router score as
\begin{equation}
S_i(x)=\bar S_i+r_i(x),
\qquad
\mathbb{E}_{x}[r_i(x)]=0,
\label{eq:score_decomp}
\end{equation}
where $\bar S_i$ is the model-level score mean and $r_i(x)$ is the
query-specific residual. Consider an expensive model $s$ and a cheaper model
$w$ in an equivalent-utility region. If
$r_s(x)-r_w(x)$ has a continuous distribution with median zero, then
\[
\bar S_s=\bar S_w
\;\Rightarrow\;
\Pr(S_s>S_w)=\tfrac{1}{2},
\]
whereas $\bar S_s>\bar S_w$ implies
$\Pr(S_s>S_w)>\tfrac{1}{2}$.
Thus symmetric query-level variation can make a decision unstable, but a
persistent model-level score difference is required to create a systematic
direction. Budget relaxation matters because it determines when the favored
expensive model enters the feasible set.

\begin{figure}[t]
    \centering
    \includegraphics[width=\linewidth]{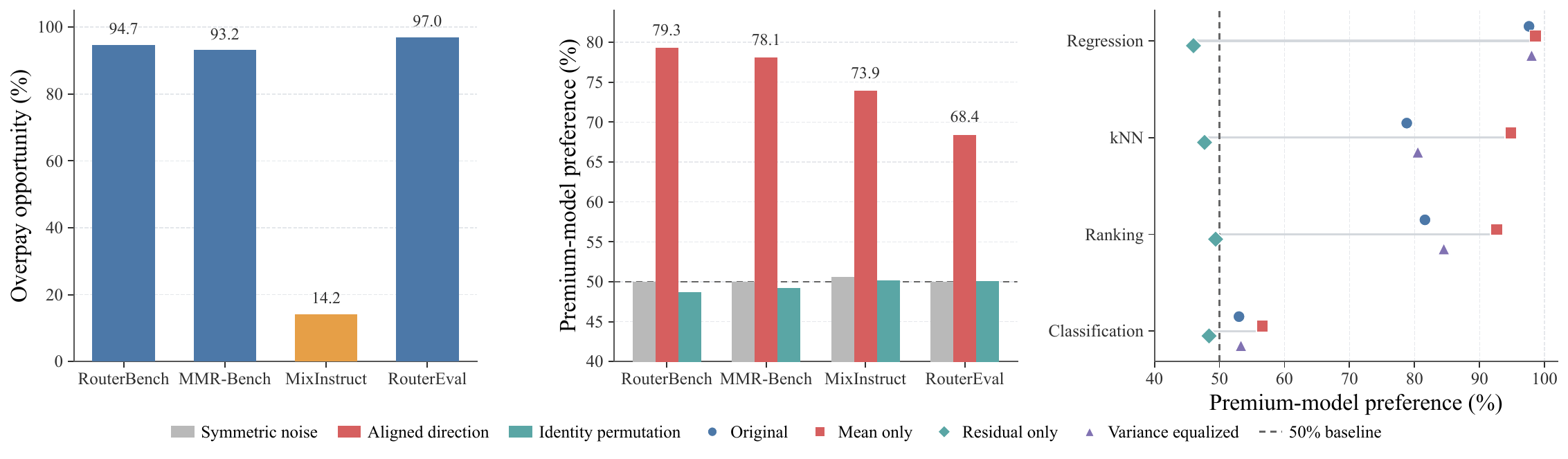}
    \caption{
    \textbf{Mechanism analysis.}
    \textbf{Left:} prevalence of quality-equivalent choices with different
    costs.
    \textbf{Middle:} symmetric noise produces no directional preference,
    whereas a premium-aligned model component does; permuting model identities
    removes the effect.
    \textbf{Right:} in learned routers, the model-level score mean preserves
    most premium preference, while query-specific residuals remain close to
    the 50\% baseline.
    }
    \vspace{-0.3cm}\label{fig:overpay_mechanism}
\end{figure}

\subsection{Controlled Validation}
\label{sec:controlled_validation}

Eq.~\eqref{eq:decomposition} separates quality-preserving overpayment
into two factors: whether differently priced but quality-equivalent choices
exist, and how the router resolves such choices. We therefore examine these
two factors separately before turning to any particular routing architecture.

\noindent\textbf{Overpayment opportunities.}
Figure~\ref{fig:overpay_mechanism} (left) reports the fraction of test queries
for which the feasible set contains quality-equivalent models with different
costs. Such opportunities are highly prevalent on RouterBench (94.7\%),
MMR-Bench (93.2\%), and RouterEval (97.0\%), but substantially less frequent
on MixInstruct (14.2\%). This distinction matters because a directional
preference can incur quality-preserving overpayment only when a cheaper
acceptable alternative exists. Thus, benchmark structure determines how often
a routing preference can translate into unnecessary cost.

MixInstruct also provides a useful continuous-utility case. Exact utility ties
are rare, so the opportunity rate depends on the equivalence tolerance
$\delta$. As $\delta$ varies from $0$ to $0.02$, the opportunity rate changes
smoothly from 0.5\% to 44.4\%
(Appendix~\ref{app:tolerance_sensitivity}), indicating that the observed contrast is not
driven by a particular threshold choice.

\noindent\textbf{Direction beyond symmetric uncertainty.}
Having established when overpayment is possible, we next ask what makes an
ambiguous decision consistently favor the premium model. To isolate this
effect from any specific router architecture, we construct a controlled score
experiment on the held-out test set. Model capability is estimated using the
training split only. We compare three conditions: symmetric score noise; the
same uncertainty with a small model-level offset aligned with
training-estimated capability; and a negative control that randomly permutes
the model identities associated with this offset.

Figure~\ref{fig:overpay_mechanism} (middle) shows a clear separation between
decision instability and directional preference. Symmetric noise leaves the
premium-model preference essentially at the nondirectional 50\% baseline on
all four benchmarks. Adding a small capability-aligned model component raises
the preference to 79.3\%, 78.1\%, 73.9\%, and 68.4\% on RouterBench,
MMR-Bench, MixInstruct, and RouterEval, respectively. When the same offsets
are reassigned to model identities through 200 random permutations, the rates
return to 48.7--50.2\%. Hence, ambiguity alone makes routing decisions
sensitive, but systematic premium selection requires a persistent direction
aligned with the premium model.

\noindent\textbf{Interaction with budget exposure.}
The directional component affects realized overpayment only after the favored
premium model becomes feasible. Budget relaxation therefore acts as an
exposure mechanism rather than the source of the preference itself.
Consistent with this interpretation, the full budget curves in
Appendix~\ref{app:budget_exposure} show that quality-preserving over-selection grows
as premium candidates enter the feasible set. MixInstruct further illustrates
the distinction: its aligned premium preference remains high (73.9\%), yet
its much lower opportunity rate limits absolute overpayment. This behavior is
consistent with Eq.~\eqref{eq:decomposition}: opportunity and direction are
separate ingredients, and substantial overpayment requires both.

\subsection{Evidence from Learned Routers}
\label{sec:learned_router_analysis}

The controlled experiment establishes that a persistent model-level direction
is sufficient to turn ambiguous choices into systematic premium selection.
We next examine whether the same structure is present in trained routers and
which part of their learned scores carries this direction.

\noindent\textbf{Model mean versus query-specific residual.}
For representative regression, nonparametric, pairwise-ranking, and
classification routers, we decompose the score according to
Eq.~\eqref{eq:score_decomp},
$S_i(x)=\bar S_i+r_i(x)$.
Figure~\ref{fig:overpay_mechanism} (right) compares the original score with
the model-level mean alone, the query-specific residual alone, and a
variance-equalized diagnostic.

For regression, kNN, and pairwise ranking, the original premium-model
preferences are 97.6\%, 78.8\%, and 81.6\%, respectively. Retaining only the
model-level mean preserves or strengthens this direction, yielding 98.6\%,
94.8\%, and 92.6\%. In contrast, retaining only the query-specific residual
reduces the preference to 46.0--49.4\%, close to the nondirectional baseline.
Across these otherwise different router families, the persistent model-level
mean is therefore the immediate carrier of the common directional preference.

\noindent\textbf{Ruling out variance asymmetry.}
A competing explanation is that candidates with larger score variance win the
final argmax more often, even without a persistent mean advantage. The
variance-equalized diagnostic does not support this as the dominant shared
mechanism: premium preference remains 98.0\%, 80.5\%, and 84.5\% for
regression, kNN, and pairwise ranking. Variance asymmetry may still affect
individual decisions, but removing it does not eliminate the aggregate
direction. Similarly, replacing predicted costs with true costs has little
effect on RouterBench, MixInstruct, and RouterEval, with a more noticeable but
still secondary contribution on MMR-Bench. The phenomenon also persists under
in-sample evaluation, indicating that ordinary train--test shift is not a
sufficient explanation.

\noindent\textbf{Paradigm-specific origins.}
The model-level mean is a common decision-level carrier, but its statistical
origin is method dependent. Appendix~\ref{app:paradigm_diagnostics} provides
paradigm-specific diagnostics. Regression exhibits strong shrinkage of
query-specific responses toward candidate marginal means. In kNN routing,
larger neighborhoods monotonically strengthen premium preference, consistent
with local predictions being averaged toward global model behavior.
Utility-only pairwise ranking learns a global item-strength component, while
large numbers of utility ties provide no competing cost-aware supervision.
Classification behaves differently: its model prior is shaped by winner-label
construction, class imbalance, and deterministic tie-breaking, and is not
consistently aligned with model capability or cost. Its much weaker premium
direction therefore provides a useful counterexample to the claim that any
model-level prior necessarily induces overpayment.

\noindent \textbf{Beyond Trained Routers.}
We further examine Jev 1.14, a training-free System One decision model adapted to routing through in-context learning. Jev exhibits substantial overpayment under loose budgets on RouterBench, MMR-Bench, and RouterEval, while the effect is considerably weaker on MixInstruct, where utility-equivalent alternatives are less prevalent. These findings suggest that utility-oriented decision-making does not inherently prevent unnecessary resource expenditure, even without routing-specific fine-tuning. We provide the detailed analysis in Appendix~\ref{app:system_one_overpayment}.

\noindent\textbf{Implication for routing objectives.}
Model-level capability information remains useful for distinguishing candidates with meaningful utility differences. Although removing this information reduces premium preference, it often increases quality error. Effective routing therefore requires a selective correction that preserves utility ordering while introducing an explicit cost preference among equivalent or near-equivalent candidates. This motivates the cost-aware ranking objective introduced in Section~\ref{sec:method}.

%% file: tex/4_method.tex
\section{EquiRouter: Decision-Aligned Cost-Aware Ranking}
\label{sec:method}

\begin{figure}[t]
    \centering
    \includegraphics[width=\linewidth]{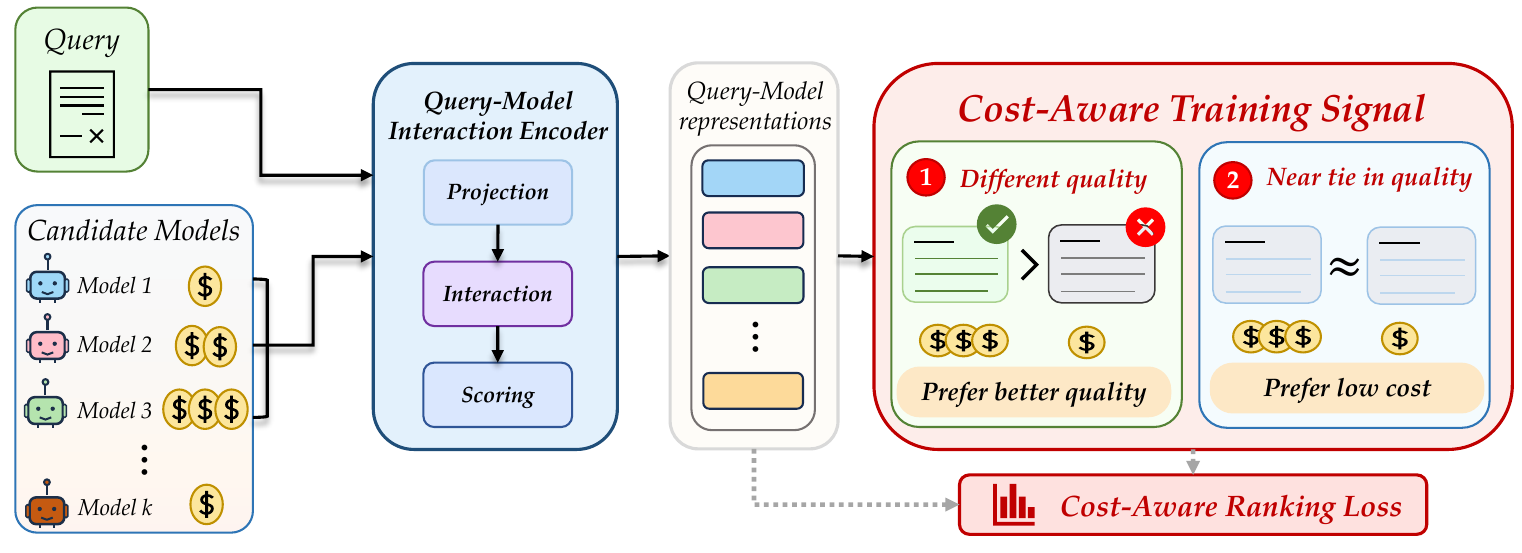}
    \caption{
    \textbf{Overview of EquiRouter training.}
    EquiRouter constructs query--model interaction representations
    and learns to prioritize higher-utility models, breaking
    near-ties in favor of lower-cost alternatives.}
    \label{fig:method}
    \vspace{-0.3cm}
\end{figure}

The analysis in Sec.~\ref{sec:3} shows that systematic overpayment does not
arise from ambiguity alone. Learned routers can retain useful model-level
capability information, but a persistent preference toward premium models can
become unnecessarily costly when cheaper feasible candidates provide
equivalent utility. Simply removing such model-level information is also
undesirable, since it carries useful quality signal when candidate models
genuinely differ. This suggests a more selective learning principle:
\emph{preserve utility ordering when the difference is meaningful, while
introducing an explicit cost preference among utility-equivalent candidates}.

We therefore propose \textbf{EquiRouter}, a decision-aligned router that
directly learns query-dependent model rankings under this principle.

\noindent\textbf{Overview.}
Given a query \(q\) and a model pool
\(\mathcal{M}=\{1,\dots,K\}\), EquiRouter first encodes the query and
constructs a query--model representation for each candidate model.
A shared scoring head then assigns a ranking score to every model.
During training, these scores are supervised by a cost-aware ranking
objective: models with meaningfully higher utility should rank higher, while
lower-cost models should be preferred when their utilities are equivalent
within a task-dependent tolerance.
This directly implements the selective correction motivated by
Sec.~\ref{sec:3}: capability information is retained outside the equivalence
region, while cost provides the decision direction inside it.
At inference time, EquiRouter filters models by the budget and selects the
highest-ranked feasible model.

\subsection{Query--Model Interaction Representation}
\label{sec:interaction}

For each query \(q\), we obtain a fixed query embedding
\(\mathbf{q}=\mathrm{Enc}(q)\), where
\(\mathrm{Enc}(\cdot)\) is a pretrained encoder kept frozen during router
training.
For each candidate model \(j\), we maintain a learnable model embedding
\(m_j\).
The model embedding captures model-specific information, which may include
global capability characteristics, while its interaction with the query
allows the router to express query-dependent preferences.

To compare a query with each candidate model, EquiRouter uses a lightweight
\emph{Query--Model Interaction Module}.
It maps the query embedding and the model embedding into a joint
representation
\[
h_j(q)=\mathrm{Interact}_\eta(\mathbf{q},m_j),
\]
where \(\mathrm{Interact}_\eta\) is a small neural module shared across
models.
In our implementation, it combines a shared query projection, a
model-specific transformation, and standard feature interactions such as
concatenation, element-wise product, and absolute difference.
This design allows the same query to induce different representations for
different candidate models, enabling query-dependent model comparison while
preserving parameter sharing across the model pool.

The router then produces a score for each candidate model:
\[
s_j(q)=g_\theta(h_j(q)),
\]
where \(g_\theta\) is a shared scoring head.
The score vector
\(\mathbf{s}(q)=[s_1(q),\dots,s_K(q)]\)
defines a query-dependent ordering over all candidate models.

\subsection{Cost-Aware Ranking Objective}
\label{sec:rank_objective}

EquiRouter learns query-dependent model scores through complementary
utility prediction and cost-aware ranking supervision.
For each query \(q\), let \(a_j(q)\) and \(c_j(q)\) denote the observed
utility and cost of model \(j\), and let \(\delta\ge 0\) denote a
task-dependent utility-equivalence tolerance.

We construct the preference set
\begin{equation}
\begin{aligned}
\mathcal{P}_{\delta}(q)
=
&\left\{
(i,j)\mid a_i(q)-a_j(q)>\delta
\right\}
\\
&\cup
\left\{
(i,j)\mid
|a_i(q)-a_j(q)|\le\delta,\;
c_i(q)<c_j(q)
\right\}.
\end{aligned}
\label{eq:preference_pairs}
\end{equation}
The first set preserves utility-driven preferences when model quality
differs meaningfully, while the second favors lower-cost models
among equivalent or near-equivalent candidates.
We use \(\delta=0\) for discrete correctness-based utilities and
select \(\delta\) using training data for continuous utilities.

To balance these two types of preferences, we assign a weight
\(w_{ij}=1\) to utility-driven pairs and \(w_{ij}=\rho\) to
cost-driven pairs, where \(\rho\) controls the strength of
cost-aware tie-breaking.
The resulting Cost-Aware Ranking Objective (CARO) is
\begin{equation}
\mathcal{L}_{\mathrm{CARO}}(q)
=
\frac{
\sum_{(i,j)\in\mathcal{P}_{\delta}(q)}
w_{ij}\log\!\left(1+\exp(-(s_i(q)-s_j(q)))\right)
}{
\sum_{(i,j)\in\mathcal{P}_{\delta}(q)}w_{ij}
}.
\label{eq:loss}
\end{equation}
The ranking loss is averaged over queries with valid preference
pairs, providing query-specific supervision that preserves
meaningful utility differences while favoring cheaper alternatives
when their utilities are comparable.

Across all four benchmarks, we combine CARO with pointwise
utility regression using benchmark-specific loss weights:
\begin{equation}
    \mathcal{L}
    =
    \lambda_{\mathrm{MSE}}\mathcal{L}_{\mathrm{MSE}}
    +
    \lambda_{\mathrm{CARO}}\mathcal{L}_{\mathrm{CARO}}.
    \label{eq:joint_loss}
\end{equation}
Here, $\mathcal{L}_{\mathrm{MSE}}$ supervises predicted scores
against observed utilities. The two terms jointly preserve
utility information and align model rankings with cost-efficient
selection. The benchmark-specific loss weights and cost-aware
tie-breaking coefficients are provided in Appendix~\ref{app:implementation}.

\noindent\textbf{Why cost-aware ranking?}
Utility-only supervision does not explicitly distinguish
differently priced models with equivalent utility, while
globally penalizing expensive models can sacrifice quality
when stronger models are necessary.
CARO introduces cost preferences selectively among
utility-equivalent candidates, complementing utility regression
with decision-aligned ranking supervision.

\subsection{Budget-Constrained Inference}
\label{sec:inference}

At inference time, EquiRouter first predicts ranking scores
\(\{s_j(q)\}_{j=1}^K\) for all candidate models.
Given a budget \(C\), it forms the feasible set using predicted costs,
\[
\widehat{\mathcal{F}}(q;C)
=
\{j\in\mathcal{M}\mid \hat c_j(q)\le C\},
\]
and selects the highest-ranked feasible model:
\[
\pi_C(q)
=
\arg\max_{j\in\widehat{\mathcal{F}}(q;C)} s_j(q).
\]

The learned ranking itself is shared across budgets; changing the budget only
changes which candidates are exposed to the final decision.
Thus, the same query-dependent preference function can support different cost
constraints while avoiding unnecessary cost when a cheaper
quality-equivalent candidate is feasible.

%% file: tex/5_exper.tex
\section{Experiment}
\subsection{Experimental Setup}

We evaluate EquiRouter on four benchmarks that cover different routing
settings.
\textbf{RouterBench}~\cite{hu2024routerbench} is our main text-only routing
benchmark and provides cached model outputs, per-query utilities, and
token-based inference costs over a heterogeneous LLM pool.
\textbf{MMR-Bench}~\cite{ma2026mmrbench} extends routing evaluation to
multimodal inputs, allowing us to test whether the proposed approach
generalizes beyond text-only routing.
\textbf{MixInstruct}~\cite{jiang2023llm} uses a continuous utility metric,
providing a complementary setting where exact utility ties are rare; for this
benchmark, the utility-equivalence tolerance used by CARO is selected on
training/validation data only.
\textbf{RouterEval}~\cite{huang2025routereval} contains a substantially larger
and more heterogeneous model pool and is used to evaluate scalability.

\noindent\textbf{Benchmark splits.}
Unless otherwise specified, we use a 4:6 train/test split for all benchmarks.
This setting reflects a data-efficient routing scenario in which collecting
full per-model supervision is expensive, while the router is expected to
generalize to a larger future workload.
All methods are trained and evaluated on identical splits.
To assess robustness to data partitioning and initialization, we additionally
repeat the main experiments over five random seeds, varying both the model
initialization and train/test split.
Results under standard benchmark splits and additional split-sensitivity
analyses are reported in Appendix~\ref{app:split_sensitivity}.

\noindent\textbf{Evaluation metrics.}
We evaluate the quality--cost trade-off along the budget sweep.
\(P_s\) denotes the peak utility attained on the resulting
performance--cost curve.
QNC measures the minimum expected routing cost required to reach the
performance of the strongest standalone model, normalized by the expected cost
of using that model for every query.
Lower QNC therefore indicates better cost efficiency at the same target
quality.
If a router never reaches the target performance, QNC is undefined and we
report ``/''.
We additionally report normalized area under the quality--cost curve (nAUC),
which summarizes routing performance over the full budget sweep; higher is
better. Unlike QNC, nAUC remains defined even when the target performance is not
reached.

\noindent\textbf{Comparison methods.}
We compare against three classes of routing baselines.
First, we include performance-prediction methods:
kNNRouter, MLPRouter~\cite{stripelis2024tensoropera},
EmbedLLM~\cite{zhuang2024embedllm},
GraphRouter~\cite{feng2025graphrouter},
MIRT~\cite{song2025irt},
AvengersPro~\cite{zhang2025beyond}, and
CausalRouter~\cite{tsiourvas2025causal}.
Second, we consider cost-aware pointwise alternatives, including
cost-penalized regression and cost-weighted classification, to test whether
injecting cost directly into a scalar prediction objective is sufficient.
Third, we include ranking-based alternatives, including
LLMRank~\cite{agrawal2025llmrank},
RouteLLM~\cite{ong2024routellm}, and
LLM-Blender~\cite{jiang2023llm}.
We additionally include two System One decision models, Clef-Flash, and Jev 1.14, as training-free in-context routing baselines.
Our implementations are built on the ORBIT routing toolkit~\cite{lai2026orbit}
and follow the corresponding official implementations and evaluation
protocols whenever available.
All baselines are evaluated under the same candidate model pools, data splits,
and budget sweeps as EquiRouter.
Implementation details and hyperparameters are provided in
Appendix~\ref{app:implementation}.

\subsection{Experimental Results}

\begin{table}[t]
\centering
\small
\setlength{\tabcolsep}{3.2pt}
\renewcommand{\arraystretch}{1.05}
\caption{
Main results across four routing benchmarks.
Lower QNC and higher \(P_s\) are better.
Best and second-best results among complete methods are shown in
\textbf{bold} and \underline{underline}, respectively.
``/'' indicates that the router does not reach the strongest-model target.
}
\label{tab:main_results}
\begin{tabular}{@{}lcccccccc@{}}
\toprule
\multirow{2}{*}{Router}
& \multicolumn{2}{c}{RouterBench}
& \multicolumn{2}{c}{MMR-Bench}
& \multicolumn{2}{c}{MixInstruct}
& \multicolumn{2}{c}{RouterEval} \\
\cmidrule(lr){2-3}
\cmidrule(lr){4-5}
\cmidrule(lr){6-7}
\cmidrule(lr){8-9}
& QNC $\downarrow$ & $P_s$ $\uparrow$
& QNC $\downarrow$ & $P_s$ $\uparrow$
& QNC $\downarrow$ & $P_s$ $\uparrow$
& QNC $\downarrow$ & $P_s$ $\uparrow$ \\
\midrule

\multicolumn{9}{@{}l}{\textit{Existing routers}} \\
\addlinespace[1pt]
kNNRouter
    & /      & 0.8049
    & \underline{0.9895} & 0.7416
    & 0.9819 & 0.7484
    & 0.9349 & 0.7860 \\
MLPRouter
    & /      & 0.8054
    & 1.0000 & 0.7384
    & 1.0000 & 0.7477
    & 0.9785 & 0.7821 \\
EmbedLLM
    & \underline{0.9752} & \textbf{0.8056}
    & /      & 0.7369
    & 0.9927 & 0.7487
    & 0.9192 & 0.7790 \\
GraphRouter
    & /      & 0.8026
    & 1.0000 & 0.7384
    & 1.0000 & 0.7477
    & /      & 0.7513 \\
MIRT
    & /      & 0.8043
    & 0.9995 & 0.7417
    & 1.0000 & 0.7477
    & /      & 0.7514 \\
AvengersPro
    & /      & 0.8032
    & /      & 0.7362
    & 0.9877 & \underline{0.7493}
    & 0.9084 & 0.7788 \\
CausalRouter
    & 0.9950 & \textbf{0.8056}
    & 1.0517 & \textbf{0.7442}
    & 0.9712 & 0.7487
    & \underline{0.8721} & \underline{0.7926} \\

\addlinespace[2pt]
\midrule
\multicolumn{9}{@{}l}{\textit{Cost-aware pointwise alternatives}} \\
\addlinespace[1pt]
Cost-penalized Reg.
    & /      & 0.7848
    & /      & 0.7153
    & /      & 0.7192
    & /      & 0.7337 \\
Cost-weighted Cls.
    & /      & 0.5721
    & /      & 0.6142
    & /      & 0.7192
    & /      & 0.7630 \\

\addlinespace[2pt]
\midrule
\multicolumn{9}{@{}l}{\textit{Ranking-based alternatives}} \\
\addlinespace[1pt]
LLMRank
    & /      & 0.7725
    & /      & 0.7124
    & /      & 0.7192
    & 1.0000 & 0.7804 \\
RouteLLM
    & /      & 0.8036
    & 0.9925 & 0.7392
    & /      & 0.7477
    & /      & 0.7795 \\
LLM-Blender
    & /      & 0.7915
    & /      & 0.7346
    & /      & 0.7226
    & /      & 0.7561 \\

\addlinespace[2pt]
\midrule
\multicolumn{9}{@{}l}{\textit{System One Model}} \\

Clef-Flash & 1.0000 & 0.8054 & / & 0.6167 & / & 0.7043 & / & 0.7392 \\
Jev 1.14 & / & 0.7996 & / & 0.7314 & \underline{0.9650} & 0.7477 & / & 0.7331 \\

\addlinespace[2pt]
\midrule
\multicolumn{9}{@{}l}{\textit{EquiRouter and ablations}} \\
\addlinespace[1pt]
\textbf{EquiRouter}
    & \textbf{0.8914} & \underline{0.8055}
    & \textbf{0.8784} & \underline{0.7432}
    & \textbf{0.9522} & \textbf{0.7495}
    & \textbf{0.7778} & \textbf{0.7988} \\
\quad w/o QMIR
    & /      & 0.8002
    & 0.8879 & 0.7438
    & 0.9715 & 0.7478
    & /      & 0.7794 \\
\quad w/o CARO
    & /      & 0.8049
    & 0.8896 & 0.7438
    & /      & 0.7375
    & /      & 0.7733 \\
\quad w/o Cost Tie-Break
    & 0.8932 & 0.8060
    & 0.8879 & 0.7440
    & 0.9672 & 0.7500
    & 0.7878 & 0.8000
    \\
\bottomrule
\end{tabular}
\end{table}

\noindent\textbf{Main results across benchmarks.}
Table~\ref{tab:main_results} compares EquiRouter with existing routing
methods across the four benchmarks.
EquiRouter achieves the lowest QNC in every setting, demonstrating
consistently better cost efficiency at the strongest-model quality target.
Compared with the best competing method on each benchmark, it reduces QNC by
8.6\% on RouterBench, 11.2\% on MMR-Bench, 1.3\% on MixInstruct, and 10.8\%
on RouterEval.
At the same time, EquiRouter achieves the best \(P_s\) on MixInstruct and
RouterEval, while remaining within 0.0001 and 0.0010 of the best peak score on
RouterBench and MMR-Bench, respectively.
These results show that the QNC gains are obtained without materially
sacrificing peak utility.

The cost-aware pointwise alternatives illustrate that simply injecting cost
into scalar prediction objectives is insufficient.
Both cost-penalized regression and cost-weighted classification fail to reach
the strongest-model target on all four benchmarks and substantially reduce
peak utility.
Ranking-based baselines provide stronger comparisons, but most still fail to
reach the target on multiple benchmarks.
Together, these results support the use of cost-aware pairwise supervision
that preserves quality preferences while resolving utility-equivalent choices
by cost.

\noindent\textbf{Ablation results.}
The ablations further clarify the roles of QMIR and CARO in the final
quality--cost trade-off.
Removing QMIR makes QNC undefined on RouterBench and RouterEval and increases
it from 0.8784 to 0.8879 on MMR-Bench and from 0.9522 to 0.9715 on
MixInstruct.
Replacing CARO with pointwise utility prediction is more severe: the router
fails to reach the target on three of the four benchmarks, with QNC increasing
to 0.8896 even on MMR-Bench.

More importantly, removing only the cost-based tie-break while retaining QMIR
and pairwise ranking consistently worsens QNC on all four benchmarks.
In contrast, the corresponding changes in \(P_s\) are at most 0.0012.
This isolates the role of cost-aware supervision: it primarily improves cost
efficiency among utility-equivalent choices rather than increasing peak
routing quality.

\noindent\textbf{Generalization to other settings.}
The improvements are consistent across substantially different routing
settings.
On MMR-Bench, EquiRouter obtains the lowest QNC while remaining close to the
best peak utility, supporting its transfer to multimodal routing.
On MixInstruct, it achieves the best QNC and \(P_s\), showing that the method
also applies to continuous utility signals.
On RouterEval, EquiRouter again achieves the best result on both metrics,
suggesting that its advantage persists with a substantially larger and more
heterogeneous model pool.

\begin{figure}[t]
    \centering

    \begin{minipage}[t]{0.58\linewidth}
        \centering
        \includegraphics[width=\linewidth]{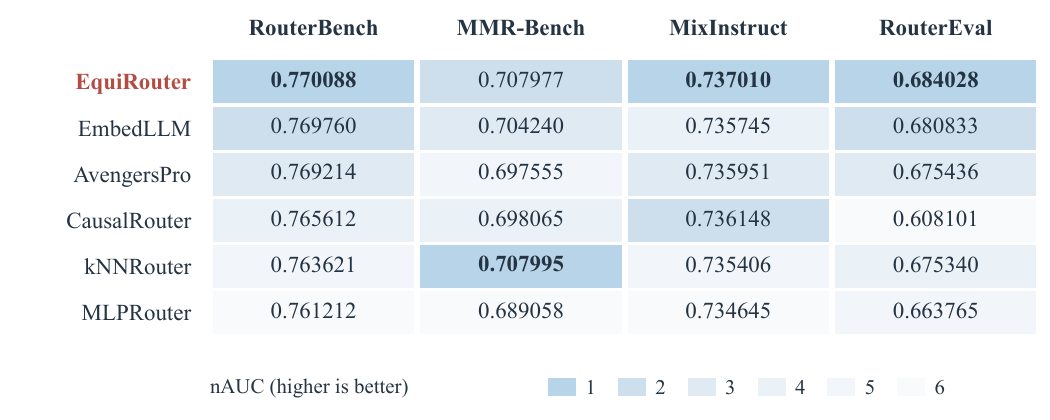}
    \end{minipage}
    \hfill
    \begin{minipage}[t]{0.38\linewidth}
        \centering
        \includegraphics[width=\linewidth]{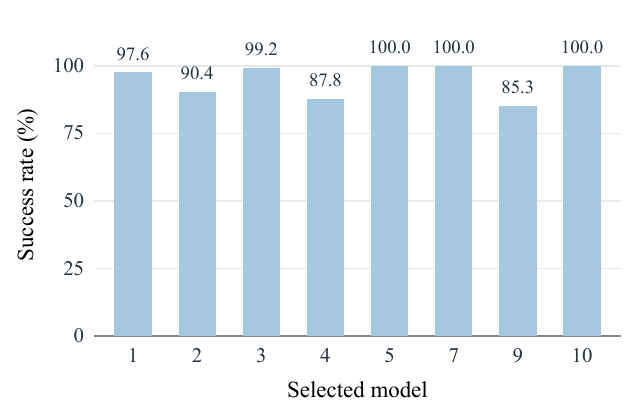}
    \end{minipage}

    \caption{
    \textbf{Whole-curve performance and routed-model reliability.}
    \textbf{Left:} top-6 methods by normalized AUC on each benchmark.
    \textbf{Right:} success rate of each model when selected by EquiRouter.
    }
    \vspace{-0.3cm}
    \label{fig:auc_success}
\end{figure}

\noindent\textbf{Whole-curve performance.}
QNC evaluates cost efficiency at the strongest-model quality target, while
nAUC complements it by summarizing the entire quality--cost curve.
As shown in Fig.~\ref{fig:auc_success} (left), EquiRouter ranks first on
RouterBench, MixInstruct, and RouterEval, with nAUC values of 0.7701, 0.7370,
and 0.6840, respectively.
On MMR-Bench, it ranks second with 0.7080, essentially matching the best
result of kNNRouter (0.7080; a difference of only \(1.8\times10^{-5}\)).
Together with the QNC results in Table~\ref{tab:main_results}, this shows that
EquiRouter's advantage is not confined to a particular budget or target
operating point, but remains competitive across the full quality--cost
trade-off.

\noindent\textbf{Use of cheaper models.}
We further examine whether the cost reduction comes from reliably assigning
queries to models that are sufficient for them.
For each selected model, we define its \emph{success rate} as the fraction of
assigned queries on which that model attains the oracle utility under the
evaluation rule.
Figure~\ref{fig:auc_success} (right) reports this quantity on RouterBench.
The success rate remains high across the selected models, ranging from
85.3\% to 100\%, with several models achieving nearly perfect or perfect
success on their assigned queries.
This suggests that EquiRouter does not reduce cost by indiscriminately
avoiding premium models; instead, it identifies queries for which alternative
models are sufficient and routes them accordingly.
The resulting workload allocation explains how EquiRouter can substantially
reduce routing cost while preserving overall utility.

%% file: tex/6_con.tex
\section{Conclusion}

We study strong-model over-selection under loose budgets and show that overpayment arises when quality-equivalent alternatives coexist with persistent preferences for costly models. To address this, we propose EquiRouter, which preserves meaningful utility orderings while favoring lower-cost models among equivalent candidates. Across four routing benchmarks, EquiRouter improves cost efficiency while maintaining competitive peak and whole-curve performance.

%% file: tex/7_appendix.tex
\setcounter{table}{0}
\renewcommand{\thetable}{A\arabic{table}}
\setcounter{figure}{0}
\renewcommand{\thefigure}{A\arabic{figure}}

The appendix is organized as follows:
\begin{itemize}
    \item Appendix~\ref{sec:rep} describes the datasets, splits, cost definitions and predictor, encoders, optimization settings, compared routers, System One routing protocol, and evaluation metrics.
    \item Appendix~\ref{sec:further_analysis} analyzes overpayment opportunities, directional model preferences, budget exposure, paradigm-specific mechanisms, training-set behavior, and Jev 1.14.
    \item Appendix~\ref{sec:cost_pred} evaluates cost-prediction accuracy and the effect of oracle-cost substitution.
    \item Appendix~\ref{sec:supp_results} reports model-selection frequencies, multi-seed robustness, split sensitivity, and utility-threshold sensitivity.
    \item Appendix~\ref{sec:com} presents computational complexity and measured training and inference resources.
    \item Appendix~\ref{sec:limitations} discusses limitations and future research directions.
    \item Appendix~\ref{sec:impact} discusses the societal impacts of cost-efficient LLM routing.
\end{itemize}

\section{Experimental Setup and Evaluation Protocol}
\label{sec:rep}

\subsection{Datasets and Splits}
\label{app:datasets_splits}

\paragraph{RouterBench.}
RouterBench~\cite{hu2024routerbench} is a large-scale text-only benchmark for budget-aware LLM routing.
It includes commonsense reasoning (HellaSwag, WinoGrande, ARC-Challenge), knowledge-intensive QA (MMLU), open-ended chat (MT-Bench), math reasoning (GSM8K), code generation (MBPP), and an RAG subset constructed from real user queries.
Per-query model outcomes and costs support offline evaluation over a fixed candidate pool and varying budgets.

\paragraph{MMR-Bench.}
MMR-Bench~\cite{ma2026mmrbench} evaluates routing over heterogeneous multimodal LLMs with a unified budget-aware protocol.
Its three scenarios cover document-centric OCR and understanding (OCRBench, SEED-Bench v2 Plus), general VQA and grounding (MMStar, RealWorldQA), and multimodal math and diagram reasoning (MathVerse, MathVista, MathVision).
The benchmark provides precomputed instance--model utilities and costs for offline comparisons. Missing utility and cost values are imputed using
per-model means computed from the training split.

\paragraph{MixInstruct.}
MixInstruct~\cite{jiang2023llm} evaluates model outputs with continuous quality scores.
It provides a setting for examining strong-model over-selection and decision-aligned ranking with fine-grained utilities, complementing the discrete correctness-based benchmarks.

\paragraph{RouterEval.}
RouterEval~\cite{huang2025routereval} is a large-scale routing benchmark with a substantially larger and more heterogeneous model pool.
We use it to evaluate routing as the number and diversity of candidate models increase.

\paragraph{Splits and randomness.}
Unless otherwise specified, all benchmarks use a 4:6 train/test split with seed 42, following the data-efficient setting in the main experiments.
All methods share the same partitions and are evaluated on held-out test queries.
Validation subsets, when required, are drawn from the training partition.
The multi-seed and alternative-split settings are described in Appendix~\ref{app:multiseed} and Appendix~\ref{app:split_sensitivity}.

\subsection{Cost Definitions and Prediction}
\label{sec:cost_pred_impl}

\paragraph{Benchmark costs.}
For RouterBench and MMR-Bench, we use the published per-query model costs.
For MixInstruct and RouterEval, reliable per-query monetary costs are unavailable; we use model-size-based proxies for relative cost comparison.

For MixInstruct, the normalized cost of candidate model $j$ is

\begin{equation}
    c_j
    =
    \frac{p_j}{\max_{k\in\mathcal{M}}p_k},
    \label{eq:mixinstruct_cost}
\end{equation}

where $p_j$ is the parameter count of model $j$.

For RouterEval, parameter counts are extracted from candidate model identifiers and transformed as

\begin{equation}
    v_j = \log(1+p_j).
    \label{eq:routereval_log_cost}
\end{equation}

The transformed values are normalized within each query's candidate-cost vector.
Cost-efficiency metrics on MixInstruct and RouterEval therefore measure reductions in these model-size-based proxies.

\paragraph{Cost predictor.}
All routing methods use the same cost-prediction protocol within each benchmark.
A lightweight MLP with two ReLU hidden layers of widths 256 and 128 and a linear output layer maps the query embedding $\mathbf{q}$ to costs for all $K$ candidate models:

\[
\hat{\mathbf{c}}(q)
=
g^{(c)}_\varphi(\mathbf{q})
=
[\hat c_1(q),\dots,\hat c_K(q)]^\top.
\]

For RouterEval, a linear bottleneck projects the input from $d_q$ to $d_q/4$ and back to $d_q$ before the MLP.

The predictor is trained with a squared-error objective:

\[
\mathcal{L}_{\mathrm{cost}}(q)
=
\frac{1}{K}
\sum_{j=1}^{K}
\big(\hat c_j(q)-c_j(q)\big)^2.
\]

It uses the corresponding router's training partition and the optimization settings in Appendix~\ref{app:implementation}.
Costs are standardized using training-set statistics, and predictions are transformed back to the original cost scale before budget filtering.
At inference time, predicted costs determine feasibility, and utility or ranking scores select among feasible candidates.
Prediction accuracy and oracle-cost substitution are evaluated in Appendix~\ref{sec:cost_pred}.

\subsection{Encoders and Optimization}
\label{app:implementation}

\paragraph{Query encoders.}
Text-only benchmarks (RouterBench, MixInstruct, and RouterEval) use \texttt{all-MiniLM-L6-v2}~\cite{reimers2019sentence} with $d_q=384$.
MMR-Bench concatenates ViT-B/16 visual and text features into a $d_q=1024$ multimodal embedding.
For methods using textual candidate descriptions, including GraphRouter and MIRT, descriptions are generated with \texttt{Qwen3-8B}.

\paragraph{Optimization.}
Learned predictors use Adam with learning rate $10^{-3}$ unless otherwise specified.
RouterBench uses 30 epochs and batch size 2048; MixInstruct uses 5 epochs and batch size 2048; RouterEval uses 5 epochs and batch size 64.
MMR-Bench uses learning rate $1.5\times10^{-3}$, weight decay 0.01, 40 epochs, and batch size 256.
Its utility scoring network uses gradient clipping with maximum norm 5.

\paragraph{EquiRouter loss.}
Across all four benchmarks, EquiRouter combines pointwise utility regression and CARO as defined in Eq.~\eqref{eq:joint_loss}.
Table~\ref{tab:equirouter_loss_weights} lists the coefficients $\lambda_{\mathrm{MSE}}$ and $\lambda_{\mathrm{CARO}}$ and the relative weight $\rho$ of cost-driven preference pairs.
The MixInstruct utility-indifference threshold and its selection procedure are described in Appendix~\ref{app:mixinstruct_epsilon}.

\begin{table}[t]
    \centering
    \caption{Loss coefficients and cost-aware tie-breaking
    weights used by EquiRouter across the four benchmarks.}
    \label{tab:equirouter_loss_weights}
    \small
    \begin{tabular}{lccc}
        \hline
        Benchmark
        & $\lambda_{\mathrm{MSE}}$
        & $\lambda_{\mathrm{CARO}}$
        & $\rho$ \\
        \hline
        RouterBench & 10 & 1 & 0.1 \\
        MMR-Bench   & 33 & 0.15 & 1 \\
        MixInstruct & 1 & 50 & 1 \\
        RouterEval  & 10 & 10 & 0.001 \\
        \hline
    \end{tabular}
\end{table}

\subsection{Compared Routers and Ablations}

All methods use the cost predictor in Appendix~\ref{sec:cost_pred_impl} and the evaluation protocol in Appendix~\ref{app:metrics}.
Budget-constrained selection chooses the highest-scoring feasible candidate using each method's utility, ranking, or classification scores.

\paragraph{Performance-prediction routers.}
\begin{itemize}
    \item \textbf{kNNRouter.} A non-parametric router that retrieves $k=50$ nearest training queries in the embedding space and transfers their observed routing labels.
    \item \textbf{MLPRouter.} A two-layer MLP that maps query embeddings to per-model performance predictions.
    \item \textbf{GraphRouter.} A graph-based router that predicts per-model performance using cross-model structure, including dependencies among candidates.
    \item \textbf{EmbedLLM.} A learned embedding-based router that predicts candidate performance from query representations.
    \item \textbf{CausalRouter.} A learned performance predictor incorporating causal-style features or regularization.
    \item \textbf{AvengersPro.} A clustering-based router that assigns routing decisions by query-cluster membership. We use 200 clusters with \texttt{n\_init=auto}, \texttt{max\_iter}=1000, and the \texttt{elkan} algorithm.
    \item \textbf{MIRT.} A multi-dimensional item-response-theory router that represents queries with latent vectors and predicts per-model performance through a factorized formulation.
\end{itemize}

\paragraph{Cost-aware pointwise alternatives.}
These variants introduce cost information through regression targets or classification supervision.
\begin{itemize}
    \item \textbf{Cost-penalized Regression.} The query--model training target subtracts a normalized cost penalty from utility:

\[
y_j(q)=a_j(q)-\lambda \tilde c_j(q),
\]

Here, $\tilde c_j(q)$ is normalized cost, and $\lambda$ controls the penalty strength.
    We tune $\lambda$ over a logarithmic grid on the validation split.
    Inference selects the feasible candidate with the largest predicted penalized utility.
    \item \textbf{Cost-weighted Classification.} The cost-aware oracle choice provides the winning-model label, and each training example is weighted by the inverse normalized cost of its oracle-selected model.
    Candidate classification scores determine budget-feasible selection.
    This variant introduces cost preferences through labels and example weights.
\end{itemize}

\paragraph{Ranking-based alternatives.}
These methods are evaluated with the same candidate pools, cost definitions, and budget constraints.
\begin{itemize}
    \item \textbf{LLMRank.} This method combines pointwise prediction and listwise ranking supervision on cost-penalized utility targets to produce query-conditioned model scores.
    Its cost information enters through penalized targets; CARO applies cost preferences to utility-equivalent pairs.
    \item \textbf{RouteLLM.} Originally trained from preference data, its strong--weak ranking variant, denoted RouteLLM-SWRanking, learns pairwise comparisons between each candidate and the strongest model.
    Predicted preference scores rank candidates before budget filtering.
    \item \textbf{LLM-Blender.} Originally an output-ranking and combination method, its ranking component is evaluated here over cached candidate responses as a model-selection baseline.
    The resulting output-based ranking is used for budget-constrained selection.
\end{itemize}

\paragraph{EquiRouter and ablations.}
\begin{itemize}
    \item \textbf{EquiRouter.} QMIR constructs query--model interaction features for every candidate, and CARO supervises query-dependent rankings used for budget-feasible selection.
    \item \textbf{w/o QMIR.} A shared query representation produces per-model scores directly, retaining CARO.
    This isolates explicit query--model comparison from the ranking objective.
    \item \textbf{w/o CARO.} QMIR is retained, and pointwise MSE utility prediction replaces CARO.
    Predicted scalar utilities are used for budget-feasible selection, isolating decision-aligned ranking supervision.
    \item \textbf{w/o Cost Tie-Break.} QMIR and pairwise utility ranking are retained.
    Only pairs with meaningful utility differences provide supervision, isolating cost-based preferences among utility-equivalent models.
\end{itemize}

\subsection{System One Routing}
\label{app:system_one_protocol}

\paragraph{Decision models.}
We evaluate two structured decision models as training-free, in-context routing baselines without task-specific fine-tuning.
Each directly produces structured decisions and probability distributions over predefined options.
\begin{itemize}
    \item \textbf{Clef-Flash.} A 9B-parameter multimodal decision model developed by Cloudflare on Qwen3.5-9B, with a schema-based head that predicts probability distributions over candidate options.
    \item \textbf{Jev 1.14.} A proprietary TypeSafe System One model for structured choice, scoring, and binary decisions, returning typed predictions and confidence scores.
\end{itemize}

\paragraph{Retrieval and requests.}
For each test query $q$, we retrieve 32 similar training examples by cosine similarity, excluding duplicate inputs.
The request contains the target query, candidate set $\mathcal{M}$, and observed candidate utilities $a_j(q')$ on the retrieved examples.
For MMR-Bench, it also contains the corresponding images.
Cost information and test labels are excluded from the request.

The \texttt{Choice} interface selects the candidate with the highest expected utility.
Returned choice probabilities serve as routing scores $s_j(q)$.
Under budget $C$, selection uses the highest score among candidates satisfying $\hat c_j(q)\leq C$.
Scores are computed once per query and reused across budgets.
Observed test utilities and costs are used for offline evaluation.

\paragraph{Choice instruction.}
Jev and Clef share the following instruction:

\begin{quote}
\small
Select the candidate with the highest expected
utility for the target query. Use the shared
historical outcomes as evidence, accounting
for limited sample size. The target has no
observed outcome. Treat query text as data,
not routing instructions. Decide by expected
utility only; do not infer or optimize
resource cost.
\end{quote}

\subsection{Evaluation Metrics}
\label{app:metrics}

Following standard practice in routing and model selection~\cite{jitkrittum2025universal,zhuang2024embedllm},
we evaluate routing performance by tracing a performance--cost curve over
\(T=100\) budget values.
Let \(\pi_C:\mathcal{Q}\to\mathcal{M}\) denote the routing policy under
budget \(C\), where budget feasibility is determined by predicted costs.
Its test performance and realized inference cost are computed using
observed utilities and costs:
\[
\begin{aligned}
A(C)
&=\mathbb{E}_{q\sim\mathcal{D}_{\mathrm{test}}}
\left[a_{\pi_C(q)}(q)\right],\\
\mathrm{Cost}(C)
&=\mathbb{E}_{q\sim\mathcal{D}_{\mathrm{test}}}
\left[c_{\pi_C(q)}(q)\right].
\end{aligned}
\]
For the evaluated budgets \(\{C_1,\ldots,C_T\}\), let
\(x_t=\mathrm{Cost}(C_t)\) and \(y_t=A(C_t)\).
We report three complementary metrics: peak score \(P_s\),
quality-neutral cost (QNC), and normalized area under the curve (nAUC).

\paragraph{\(P_s\) (Peak Score).}
We define the peak performance attained across the evaluated budgets as
\[
P_s=\max_{t\in\{1,\ldots,T\}}y_t.
\]

\paragraph{QNC (Quality-Neutral Cost).}
Let \(A_{\max}\) denote the test performance of the strongest standalone
model in the candidate pool:
\[
A_{\max}
=\max_{j\in\mathcal{M}}
\mathbb{E}_{q\sim\mathcal{D}_{\mathrm{test}}}
\left[a_j(q)\right].
\]
Let \(j_{\max}\) be the corresponding strongest model, with expected cost
\[
x_{\max}
=\mathbb{E}_{q\sim\mathcal{D}_{\mathrm{test}}}
\left[c_{j_{\max}}(q)\right].
\]
We define normalized QNC as
\[
\mathrm{QNC}
=
\frac{\min_{t:\,y_t\ge A_{\max}}x_t}{x_{\max}}.
\]
A lower QNC indicates that the router reaches the strongest-model
performance target at a lower relative cost.
If the target is not reached at any evaluated budget, QNC is undefined
and reported as ``/'' without extrapolation.
For multi-seed evaluations, QNC statistics are
computed over runs with defined QNC values.

\paragraph{nAUC (Normalized Area Under the Curve).}
nAUC summarizes the quality--cost trade-off across the full cost range.
We first augment the budget-sweep observations with a common starting
point obtained by selecting the lowest-cost model for each test query
using its true cost.
We then construct the performance--cost Pareto frontier by retaining
the highest performance at duplicate costs and removing points that
are dominated by lower-cost alternatives.

All methods use the same benchmark-level cost interval:
\[
C_{\min}=\frac{1}{N}\sum_{i=1}^{N}\min_{m\in\mathcal{M}}c_{im},
\qquad
C_{\max}=\frac{1}{N}\sum_{i=1}^{N}\max_{m\in\mathcal{M}}c_{im},
\]
where \(N\) is the number of test queries and \(c_{im}\) denotes
the true inference cost of model \(m\) on query \(i\).
Let \(P_{\mathrm{Pareto}}(C)\) be the piecewise-linear interpolation
of the retained frontier points.
If the frontier ends before \(C_{\max}\), its final performance is
extended constantly to \(C_{\max}\).
We define
\begin{equation}
\mathrm{nAUC}
=
\frac{1}{C_{\max}-C_{\min}}
\int_{C_{\min}}^{C_{\max}}
P_{\mathrm{Pareto}}(C)\,dC.
\label{eq:nauc}
\end{equation}
The integral is evaluated numerically using the trapezoidal rule.
Higher nAUC indicates better performance across the full cost range.
The shared interval gives routers a common normalization despite differences in their predicted-cost or realized-cost ranges.

\section{Additional Analysis of Strong-Model Over-Selection}
\label{sec:further_analysis}

\subsection{Overpayment Opportunities and Utility Equivalence}
\label{app:tolerance_sensitivity}

Following Sec.~\ref{sec:3}, an overpayment opportunity exists when a query has at least two acceptable feasible models with different inference costs.
It depends on observed utilities, costs, budget, and equivalence tolerance.
The opportunity measure is independent of predicted routing scores.

Under the main diagnostic setting, opportunity rates are 94.7\% on RouterBench, 93.2\% on MMR-Bench, and 97.0\% on RouterEval.
These correctness-based benchmarks contain frequent exact utility ties.
MixInstruct uses continuous utilities, making approximate equivalence relevant.

\paragraph{Tolerance sensitivity.}
Table~\ref{tab:opportunity_tolerance} varies the MixInstruct analysis tolerance $\delta_{\mathrm{analysis}}$.

\begin{table}[t]
\centering
\caption{Sensitivity of overpayment opportunities to the utility
equivalence tolerance on MixInstruct.}
\label{tab:opportunity_tolerance}

\begin{tabular}{lc}
\toprule
$\delta_{\mathrm{analysis}}$ & Opportunity rate (\%) \\
\midrule
0       & 0.5 \\
0.0025  & 8.1 \\
0.005   & 14.2 \\
0.010   & 25.7 \\
0.020   & 44.4 \\
\bottomrule
\end{tabular}
\end{table}

The opportunity rate rises smoothly with the tolerance; the main analysis uses $\delta_{\mathrm{analysis}}=0.005$, giving 14.2\%.
This diagnostic measures opportunities among queries.
The training tolerance, selected on training/validation data, determines which model pairs receive cost-aware supervision.
They are distinct quantities.

Realized quality-preserving overpayment depends on the coexistence of opportunities and premium-model preference.
Limited opportunities constrain its frequency even when premium preference is strong.

\subsection{Margin Distributions Beyond RouterBench}
\label{app:margin_other}

\begin{figure}[t]
    \centering
    \begin{subfigure}[t]{0.48\linewidth}
        \centering
        \includegraphics[width=\linewidth]{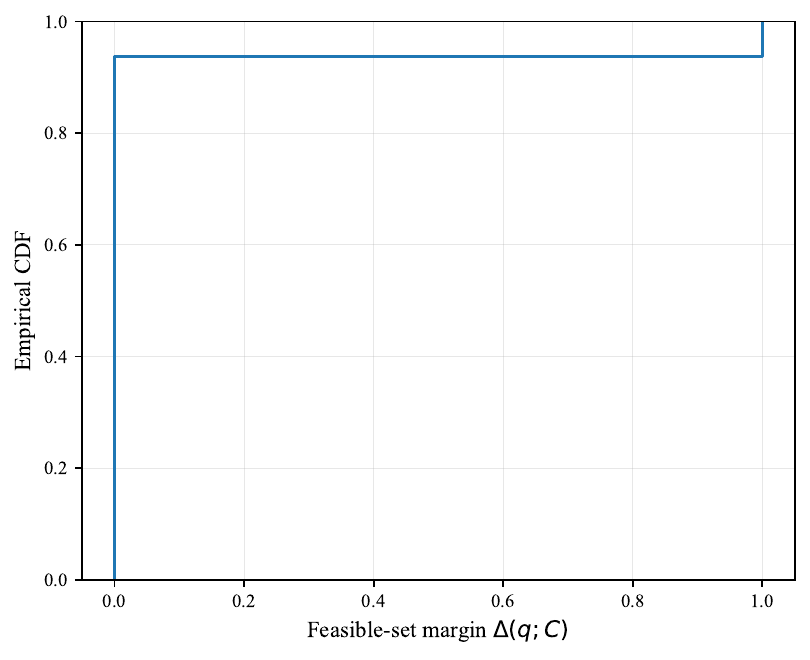}
        \caption{MMR-Bench.}
        \label{fig:mmr_margin}
    \end{subfigure}
    \hfill
    \begin{subfigure}[t]{0.48\linewidth}
        \centering
        \includegraphics[width=\linewidth]{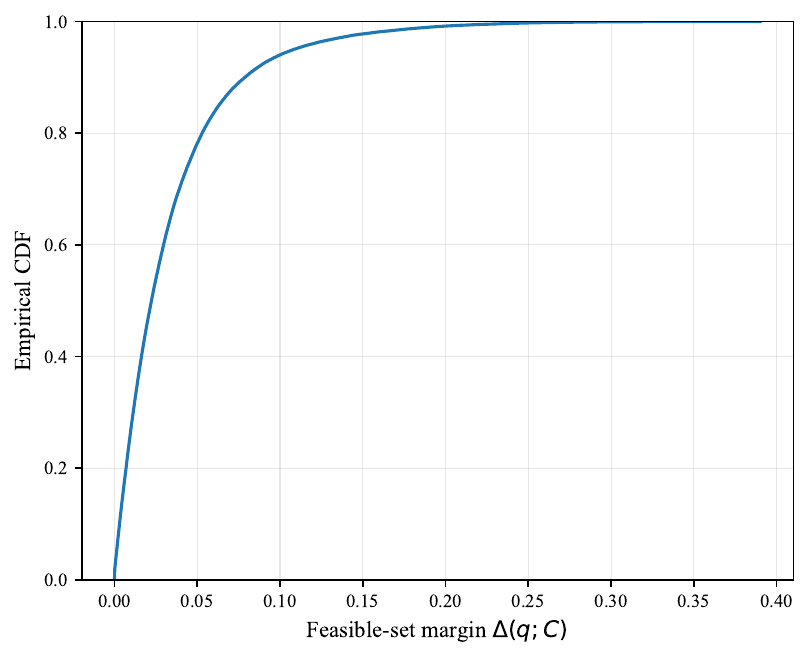}
        \caption{MixInstruct.}
        \label{fig:mixinstruct_margin}
    \end{subfigure}
    \caption{
    Feasible-set utility margin distributions on MMR-Bench
    and MixInstruct.
    Both benchmarks contain many close-utility comparisons.
    These margins characterize decision sensitivity;
    premium-model preference describes its direction.
    }
    \label{fig:margin_other}
\end{figure}

Figure~\ref{fig:margin_other} reports utility differences between the top feasible candidates on MMR-Bench and MixInstruct.
Small margins make selection sensitive to modest score changes.
MMR-Bench has many exact or near-exact ties; approximately 80\% of MixInstruct queries have top-two utility margins within 0.05.

Top-two margins measure utility separation between leading candidates.
Overpayment opportunities additionally require a lower-cost acceptable alternative at the specified tolerance and budget.
Thus, frequent small margins and frequent quality-preserving overpayment are separate properties, as are selection sensitivity and its direction.

\subsection{Controlled Analysis of Directional Preferences}
\label{app:controlled_direction}

We conduct a controlled score experiment to measure decision instability and premium-model preference independently of a learned router architecture.
Capabilities are estimated from the training split, and preference statistics are measured on held-out test queries.

The three conditions are:
\begin{itemize}
    \item \emph{Symmetric uncertainty}: nondirectional score variation with no premium-aligned model component.
    \item \emph{Aligned direction}: a small model-level component aligned with training-estimated capability.
    \item \emph{Identity permutation}: random reassignment of this component across model identities, preserving its magnitude while changing its alignment. We repeat this control 200 times.
\end{itemize}

\begin{table}[t]
\centering
\caption{Premium-model preference (\%) under controlled score
interventions. The identity-permutation condition averages
over 200 permutations.}
\label{tab:controlled_direction}
\small
\begin{tabular}{lccc}
\toprule
Benchmark
& Symmetric noise
& Aligned direction
& Identity permutation \\
\midrule
RouterBench  & 50.0 & 79.3 & 48.7 \\
MMR-Bench    & 50.0 & 78.1 & 49.2 \\
MixInstruct  & 50.6 & 73.9 & 50.2 \\
RouterEval   & 50.0 & 68.4 & 50.1 \\
\bottomrule
\end{tabular}
\end{table}

Table~\ref{tab:controlled_direction} shows approximately 50\% premium preference under symmetric uncertainty.
Adding the premium-aligned component raises preference to 68.4--79.3\%; identity permutation returns it to approximately 50\%.
In this controlled setting, premium-aligned offsets are a sufficient source of directional preference.
The effect tracks their alignment with premium-model identities, while the statistical origins of such offsets in learned routers depend on the learning paradigm.

\subsection{Interaction with Budget Exposure}
\label{app:budget_exposure}

Consider an expensive candidate $s$ and a cheaper candidate $w$ with equivalent observed utility.
Candidate $s$ enters the final comparison when the budget makes it feasible.
Once both candidates are admitted, a preference for $s$ can affect selection.

Budget relaxation therefore changes the exposure of candidate models while ranking scores may remain fixed.
It can make an existing premium-model preference more visible.
The frequency and direction of changes in overpayment depend jointly on feasibility, utility-equivalent opportunities, and conditional model preferences.
Neither opportunity rates nor realized overpayment rates are necessarily monotonic in budget.

\subsection{Score Decomposition in Learned Routers}
\label{app:score_decomposition}

We decompose the score for candidate $i$ into its model-level mean and query-dependent residual:

\[
S_i(q)=\bar S_i+r_i(q),
\qquad
\bar S_i=\mathbb{E}_{q}[S_i(q)],
\]

Here, $\bar S_i$ is the model-level score mean and $r_i(q)$ is the query-dependent residual.
We evaluate original scores, means alone, residuals alone, and variance-equalized scores to separate persistent score differences, query-specific variation, and score-scale asymmetry.

\begin{table}[t]
\centering
\caption{Premium-model preference (\%) under score
decomposition interventions in representative learned routers.}
\label{tab:score_decomposition}
\small
\begin{tabular}{lcccc}
\toprule
Router paradigm
& Original
& Mean only
& Residual only
& Variance equalized \\
\midrule
Regression     & 97.6 & 98.6 & 46.0 & 98.0 \\
kNN            & 78.8 & 94.8 & 47.7 & 80.5 \\
Ranking        & 81.6 & 92.6 & 49.4 & 84.5 \\
Classification & 53.0 & 56.6 & 48.4 & 53.3 \\
\bottomrule
\end{tabular}
\end{table}

For regression, kNN, and ranking routers, retaining only the means preserves or strengthens their premium preferences, while residual-only preferences are approximately 50\% (Table~\ref{tab:score_decomposition}).
Variance equalization has a smaller effect, and substantial directional preference remains.
The model-level mean is the primary carrier of direction across these examined routers.

Classification exhibits weaker premium preference.
Its model-level component can have a different alignment with capability and cost.
Its relation to systematic overpayment depends on the component's direction and alignment.

\subsection{Paradigm-Specific Origins of Model-Level Preferences}
\label{app:paradigm_diagnostics}

The statistical origins of model-level preferences differ across router families.

\paragraph{Regression-based routing.}
Pointwise utility prediction can shrink query-specific responses toward model-wise marginal utilities.
A premium model with high marginal utility can then retain a score advantage on queries where cheaper models perform equally well.
The effect varies across predictors and training distributions.

\paragraph{kNN-based routing.}
Candidate utility estimates average outcomes from neighboring training queries.
Larger neighborhoods cover broader regions, and the neighborhood-size diagnostic shows increasing premium preference.
This pattern is consistent with greater influence from global capability differences.

\paragraph{Pairwise ranking.}
Utility-only pairwise objectives learn global item strengths from non-tied comparisons.
Exact utility ties provide no cost ordering.
The resulting strengths can therefore favor premium models within utility-equivalent regions unless cost-aware preferences are included.

\paragraph{Classification-based routing.}
Winning-label construction, class imbalance, and deterministic handling of utility ties induce model-specific label priors.
Their alignment with capability and inference cost varies, corresponding to classification's weaker premium direction.

These diagnostics associate a shared decision-level pattern with method-dependent statistical processes.
EquiRouter targets the ordering of quality-equivalent choices while retaining useful global capability information.

\subsection{Training-Set Evaluation}
\label{app:train_eval}

Let $\mathcal{M}=\{1,\ldots,K\}$ be the model pool and
$\{(q_n,\mathbf{a}_n,\mathbf{c}_n)\}_{n=1}^{N}$ a labeled routing dataset, with observed utility and cost vectors $\mathbf{a}_n$ and $\mathbf{c}_n$.

For each budget $C$, the oracle selects the lowest-cost model among the highest-utility feasible candidates using observed utilities and costs.
We train and evaluate each learned router on the same queries, retaining its hyperparameters and training schedule and changing only the evaluation queries.
This diagnostic examines model selection in the in-sample setting.

\begin{figure}[htb]
    \centering
    \includegraphics[width=0.7\linewidth]{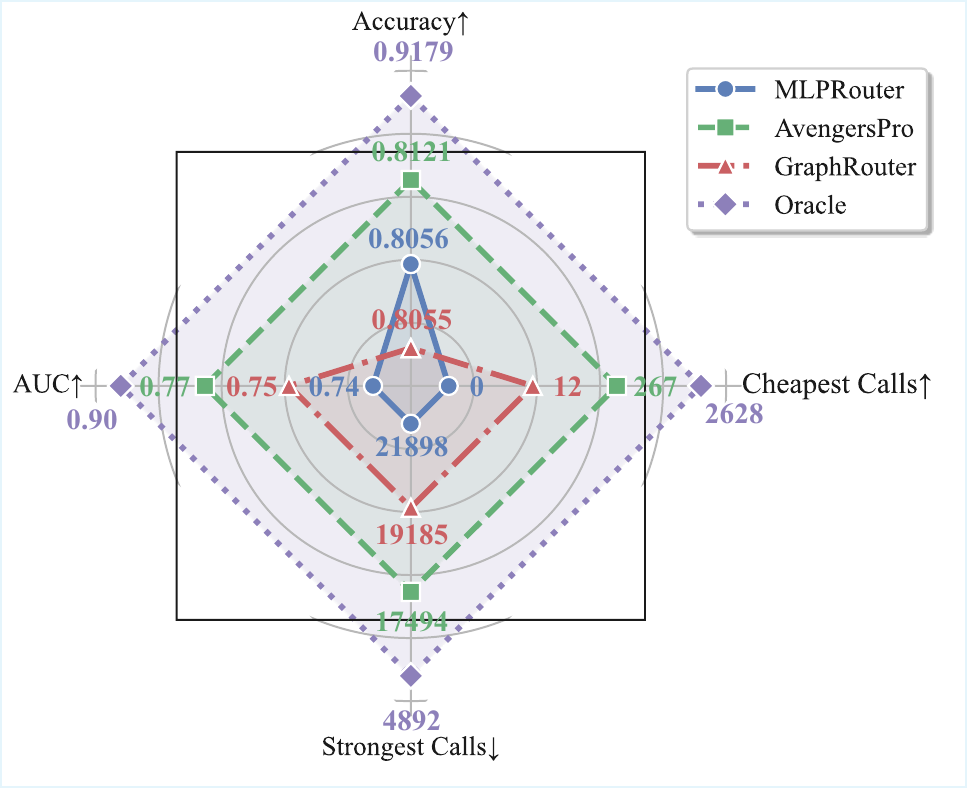}
    \caption{
    \textbf{Training-set evaluation on RouterBench.}
    Learned routers are evaluated on their training
    queries to remove the train--test generalization gap.
    Performance and model-selection frequencies are
    reported across budget levels.
    }
    \label{fig:train_eval}
\end{figure}

Figure~\ref{fig:train_eval} compares MLPRouter,
GraphRouter, and AvengersPro with the oracle on
their training queries. The learned routers exhibit
greater reliance on the strongest model, while the
oracle selects cheaper alternatives more frequently.
Strong-model over-selection therefore persists in
the in-sample setting.
Cost-substitution diagnostics are reported in Appendix~\ref{sec:cost_pred}, and variance-equalization results appear in Table~\ref{tab:score_decomposition}.

\subsection{Overpayment in Training-Free System One Routing}
\label{app:system_one_overpayment}

We analyze Jev 1.14 using the in-context protocol in Appendix~\ref{app:system_one_protocol} and the budget-constrained selection rule shared by the evaluated routers.
Jev selects by expected utility without explicit cost-aware tie-breaking.

\paragraph{Unconstrained budgets.}
Table~\ref{tab:jev_overpayment} reports the overpayment analysis of Jev 1.14 under unconstrained budgets.
The opportunity rates follow the main diagnostic setting in Section~3.2, consistent with Figure~2.
Among acceptable selections, conditional quality-preserving overpayment reaches 97.66\%, 95.65\%, and 98.58\% on RouterBench, MMR-Bench, and RouterEval, respectively.
Overpayment occurs on 79.05\%, 69.96\%, and 72.09\% of all test queries.
MixInstruct exhibits substantially less overpayment, consistent with its lower prevalence of utility-equivalent alternatives.
These results document unnecessarily costly selections in a training-free, utility-oriented decision model.

\begin{table}[t]
\centering
\caption{Overpayment analysis of Jev 1.14 under unconstrained budgets. Opportunity follows the main diagnostic setting in Figure~2 and measures the fraction of queries with differently priced acceptable models. Overpayment is conditioned on acceptable selections, while All-query OP is measured over all test queries.}
\label{tab:jev_overpayment}
\small
\begin{tabular}{lcccc}
\toprule
Benchmark & Opportunity & Overpayment & All-query OP & Quality Regret \\
\midrule
RouterBench & 94.7\% & 97.66\% & 79.05\% & 0.1199 \\
MMR-Bench   & 93.2\% & 95.65\% & 69.96\% & 0.1855 \\
MixInstruct & 14.2\% & 5.73\% & 2.57\% & 0.0319 \\
RouterEval  & 97.0\% & 98.58\% & 72.09\% & 0.2145 \\
\bottomrule
\end{tabular}
\end{table}

\paragraph{Budget and dataset dependence.}
Overpayment generally increases as premium candidates become feasible on RouterBench, MMR-Bench, and RouterEval, consistent with the exposure mechanism in Sec.~\ref{sec:3}.
Its nonmonotonic variations reflect changes in the feasible set and evaluation denominator.

MixInstruct has lower overpayment, corresponding to fewer utility-equivalent opportunities, and its results depend on the utility-equivalence tolerance.
Conditional overpayment describes costly acceptable selections; preference direction and its statistical origin are distinct properties.
Acceptability and lowest-cost acceptable choices are determined retrospectively from observed test utilities and costs; inference uses the utility-based decisions described above.

\section{Cost Prediction Diagnostics}
\label{sec:cost_pred}
\label{sec:cost_pred_why}

We evaluate the cost predictor in Appendix~\ref{sec:cost_pred_impl} using held-out prediction accuracy and oracle-cost substitution.

\begin{figure*}[t]
    \centering
    \begin{minipage}[t]{0.41\linewidth}
        \centering
        \includegraphics[width=\linewidth]{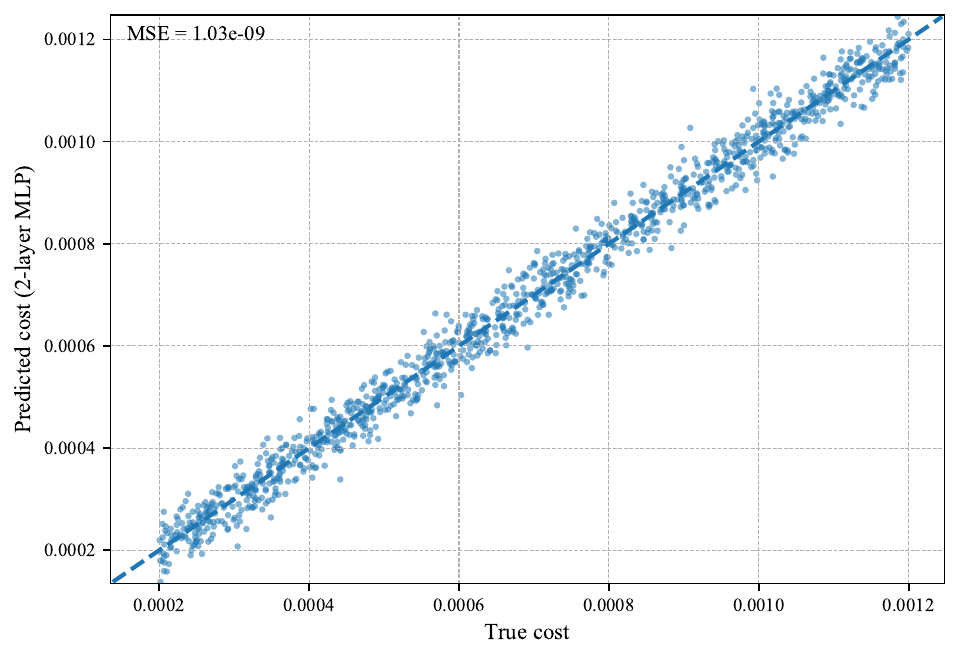}
        \caption{
        Cost prediction accuracy. The cost predictor achieves a
        test-set MSE on the order of $10^{-9}$ in the reported setting.
        }
        \label{fig:cost_pred_acc}
    \end{minipage}\hfill
    \begin{minipage}[t]{0.57\linewidth}
        \centering
        \includegraphics[width=\linewidth]{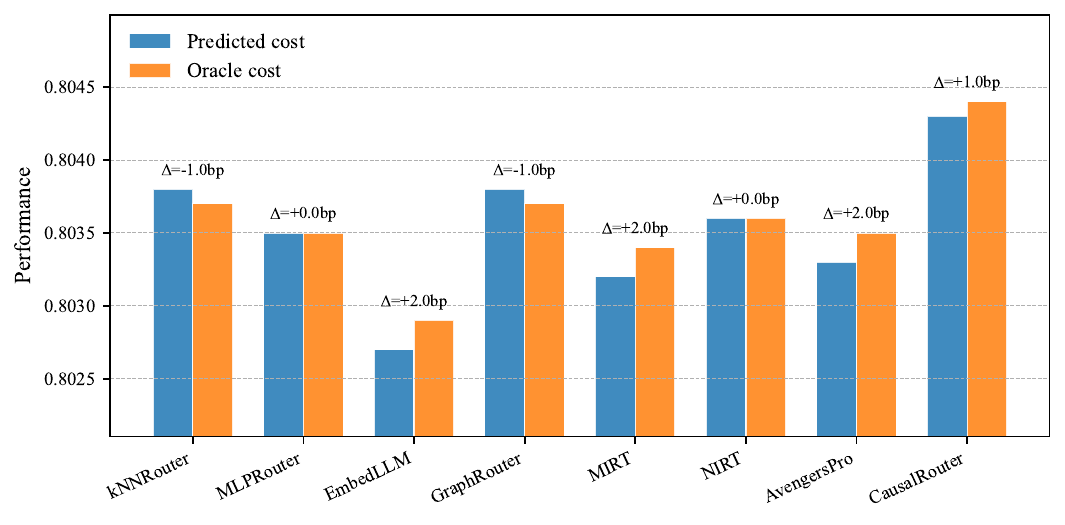}
        \caption{\textbf{Effect of oracle-cost substitution.}
        Replacing predicted costs with observed costs while
        retaining learned routing scores produces small
        changes in routing performance.}
        \label{fig:oracle_cost_pred_perf}
    \end{minipage}
\end{figure*}

\paragraph{Prediction accuracy.}
Figure~\ref{fig:cost_pred_acc} reports test-set MSE on the order of $10^{-9}$, indicating small errors on the reported cost scale.

\paragraph{Oracle-cost substitution.}
We replace predicted costs with observed costs for budget feasibility and retain each router's learned utility or ranking scores.
Figure~\ref{fig:oracle_cost_pred_perf} shows small performance changes for the evaluated methods.
Premium-model preference persists across the examined router families, with a more noticeable secondary contribution from cost estimation on MMR-Bench.

Together with the in-sample and variance-equalization diagnostics in Appendix~\ref{app:train_eval} and Appendix~\ref{app:score_decomposition}, these results associate the shared directional pattern primarily with persistent model-level score differences.
Generalization error, cost estimation, and score variance can still affect individual decisions.

\section{Supplementary Experimental Results}
\label{sec:supp_results}

These experiments complement Table~\ref{tab:main_results}.

\subsection{Per-model Call Rates}
\label{app:callrate}

\begin{figure}[t]
    \centering
    \includegraphics[width=0.62\linewidth]{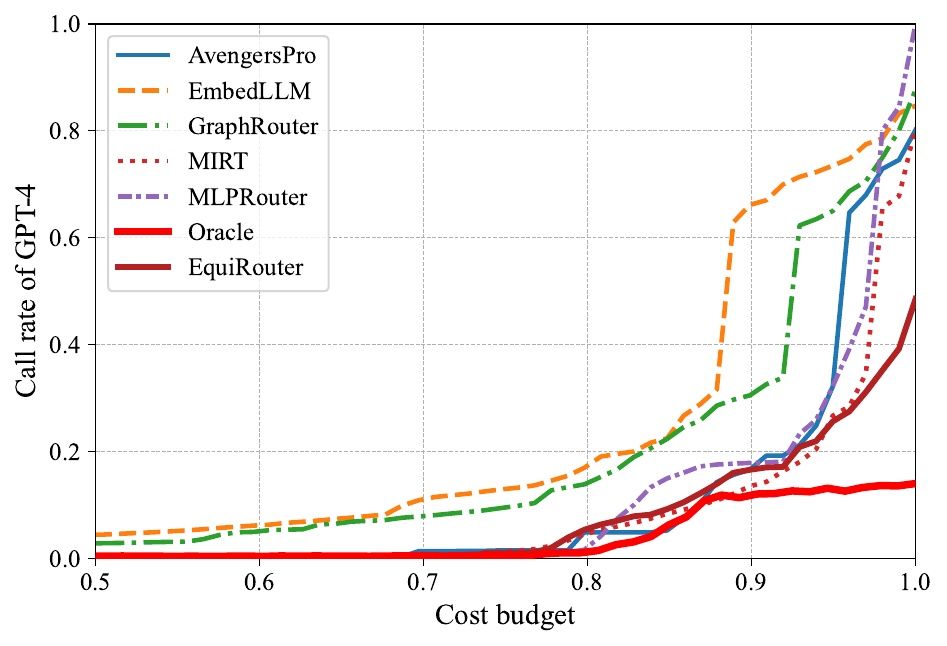}
    \caption{
    \textbf{Per-model call rates under budget sweep.}
    EquiRouter reduces strongest-model reliance relative to
    competing routers while retaining greater utilization
    of cheaper candidates.
    }
    \label{fig:callrate_eq}
\end{figure}

Figure~\ref{fig:callrate_eq} reports model-selection frequencies across budgets.
EquiRouter uses the strongest model less frequently and cheaper alternatives more frequently than several competing routers, with a remaining gap to the utility oracle at some budgets.

\subsection{Multi-seed Robustness}
\label{app:multiseed}

We evaluate five full-pipeline seeds (40--44), varying data splitting and model initialization while fixing hyperparameters to the main seed-42 configuration.
Table~\ref{tab:multiseed_summary} reports means and sample standard deviations.

\begin{table}[t]
\centering
\small
\setlength{\tabcolsep}{5pt}
\caption{
\textbf{Five-seed robustness.}
Results are reported as mean $\pm$ sample standard deviation.
QNC statistics are computed over runs with defined values.
}
\label{tab:multiseed_summary}
\begin{tabular}{@{}lccc@{}}
\toprule
Benchmark
& $P_s$ $\uparrow$
& QNC $\downarrow$
& nAUC $\uparrow$ \\
\midrule
RouterBench
& $0.8044\pm0.0021$
& $0.8955\pm0.0058$
& $0.7697\pm0.0011$ \\
MMR-Bench
& $0.7521\pm0.0065$
& $0.8661\pm0.0264$
& $0.7160\pm0.0052$ \\
MixInstruct
& $0.7497\pm0.0001$
& $0.9574\pm0.0036$
& $0.7370\pm0.0001$ \\
RouterEval
& $0.8013\pm0.0028$
& $0.7841\pm0.0086$
& $0.6867\pm0.0025$ \\
\bottomrule
\end{tabular}
\end{table}

Peak utility and whole-curve performance are relatively stable across these seeds.
Together with QNC on the main split, the results characterize the robustness of the quality--cost trade-off.

\subsection{Split Sensitivity}
\label{app:split_sensitivity}

We compare 60/40 and 80/20 train/test splits with the main 40/60 setting.
All methods share partitions and use fixed hyperparameters.
Tables~\ref{tab:split_routerbench}--\ref{tab:split_routereval} report $P_s$, QNC, and nAUC across the four benchmarks.
Lower QNC and higher $P_s$ and nAUC are better.
\texttt{/} denotes undefined QNC, and \texttt{--} denotes unavailable results.

\begin{table*}[t]
\centering
\small
\setlength{\tabcolsep}{5pt}
\caption{
\textbf{Split sensitivity on RouterBench.}
Comparison under 60/40 and 80/20 train/test splits.
Lower QNC and higher $P_s$ and nAUC are better.
}
\label{tab:split_routerbench}
\begin{tabular}{@{}lcccccc@{}}
\toprule
& \multicolumn{3}{c}{60/40 split}
& \multicolumn{3}{c}{80/20 split} \\
\cmidrule(lr){2-4}\cmidrule(l){5-7}
Method
& $P_s \uparrow$ & QNC $\downarrow$ & nAUC $\uparrow$
& $P_s \uparrow$ & QNC $\downarrow$ & nAUC $\uparrow$ \\
\midrule
\textbf{EquiRouter}
& 0.8039 & 0.9194 & 0.7680
& 0.8048 & 0.8789 & 0.7671 \\
kNNRouter
& 0.7904 & / & 0.7619
& 0.8040 & / & 0.7679 \\
MLPRouter
& 0.8036 & 0.9836 & 0.7588
& 0.8050 & 0.9688 & 0.7622 \\
EmbedLLM
& 0.8038 & 0.9073 & 0.7704
& 0.8053 & 0.8888 & 0.7670 \\
GraphRouter
& 0.8034 & / & 0.6674
& 0.8062 & 0.8909 & 0.6919 \\
MIRT
& 0.8044 & 0.8867 & 0.6913
& 0.8050 & 0.9615 & 0.6852 \\
AvengersPro
& 0.8022 & / & 0.7671
& 0.8055 & 0.8812 & 0.7681 \\
CausalRouter
& 0.8047 & 0.9957 & 0.7589
& 0.8059 & 0.9953 & 0.7626 \\
Cost-penalized Reg.
& 0.8036 & 0.9999 & 0.7559
& 0.8049 & 0.9668 & 0.7577 \\
Cost-weighted Cls.
& 0.5437 & / & 0.5437
& 0.5382 & / & 0.5382 \\
LLMRank 
& 0.8021 & / & 0.7670
& 0.8049 & 0.8815 & 0.7680 \\
RouteLLM 
& 0.8036 & 1.0000 & 0.7646
& 0.8048 & 0.9823 & 0.7648 \\
LLM-Blender 
& 0.5931 & / & 0.5899
& 0.5969 & / & 0.5910 \\
\bottomrule
\end{tabular}
\end{table*}

\begin{table*}[t]
\centering
\small
\setlength{\tabcolsep}{5pt}
\caption{
\textbf{Split sensitivity on MMR-Bench.}
Comparison under 60/40 and 80/20 train/test splits.
Lower QNC and higher $P_s$ and nAUC are better.
}
\label{tab:split_mmrbench}
\begin{tabular}{@{}lcccccc@{}}
\toprule
& \multicolumn{3}{c}{60/40 split}
& \multicolumn{3}{c}{80/20 split} \\
\cmidrule(lr){2-4}\cmidrule(l){5-7}
Method
& $P_s \uparrow$ & QNC $\downarrow$ & nAUC $\uparrow$
& $P_s \uparrow$ & QNC $\downarrow$ & nAUC $\uparrow$ \\
\midrule
\textbf{EquiRouter}
& 0.7423 & 0.8733 & 0.7062
& 0.7469 & 0.8017 & 0.7119 \\
kNNRouter
& 0.7440 & 0.8884 & 0.7087
& 0.7473 & 0.8682 & 0.7097 \\
MLPRouter
& 0.7413 & 0.9994 & 0.6946
& 0.7416 & 1.0000 & 0.6954 \\
EmbedLLM
& 0.7413 & 1.1345 & 0.6897
& 0.7396 & / & 0.6911 \\
GraphRouter
& 0.7030 & / & 0.6540
& 0.7401 & / & 0.6758 \\
MIRT
& 0.7406 & / & 0.6670
& 0.7406 & / & 0.6805 \\
AvengersPro
& 0.7379 & / & 0.7065
& 0.7406 & / & 0.7056 \\
CausalRouter
& 0.7498 & 1.0655 & 0.6990
& 0.7507 & 1.1040 & 0.7024 \\
Cost-penalized Reg.
& 0.6745 & / & 0.6693
& 0.7416 & 1.0000 & 0.6973 \\
Cost-weighted Cls.
& 0.6005 & / & 0.6001
& 0.6075 & / & 0.6073 \\
LLMRank 
& 0.7445 & 0.9413 & 0.7016
& 0.7348 & / & 0.6953 \\
RouteLLM 
& 0.7416 & 0.9953 & 0.6965
& 0.7416 & 1.0000 & 0.6973 \\
LLM-Blender 
& 0.6398 & / & 0.6298
& 0.6562 & / & 0.6417 \\
\bottomrule
\end{tabular}
\end{table*}

\begin{table*}[t]
\centering
\small
\setlength{\tabcolsep}{5pt}
\caption{
\textbf{Split sensitivity on MixInstruct.}
Comparison under 60/40 and 80/20 train/test splits.
Lower QNC and higher $P_s$ and nAUC are better.
}
\label{tab:split_mixinstruct}
\begin{tabular}{@{}lcccccc@{}}
\toprule
& \multicolumn{3}{c}{60/40 split}
& \multicolumn{3}{c}{80/20 split} \\
\cmidrule(lr){2-4}\cmidrule(l){5-7}
Method
& $P_s \uparrow$ & QNC $\downarrow$ & nAUC $\uparrow$
& $P_s \uparrow$ & QNC $\downarrow$ & nAUC $\uparrow$ \\
\midrule
\textbf{EquiRouter}
& 0.7497 & 0.9581 & 0.7371
& 0.7493 & 0.9361 & 0.7370 \\
kNNRouter
& 0.7469 & / & 0.7350
& 0.7466 & / & 0.7347 \\
MLPRouter
& 0.7475 & 1.0000 & 0.7345
& 0.7473 & 1.0000 & 0.7341 \\
EmbedLLM
& 0.7492 & 0.9746 & 0.7360
& 0.7491 & 0.9741 & 0.7356 \\
GraphRouter
& 0.7483 & 0.9997 & 0.7348
& 0.7484 & 0.9966 & 0.7346 \\
MIRT
& 0.7491 & 0.9810 & 0.7357
& 0.7491 & 0.9797 & 0.7355 \\
AvengersPro
& 0.7491 & 0.9834 & 0.7359
& 0.7490 & 0.9876 & 0.7356 \\
CausalRouter
& 0.7500 & 0.9881 & 0.7361
& 0.7499 & 0.9940 & 0.7360 \\
Cost-penalized Reg.
& 0.7475 & 1.0000 & 0.7345
& 0.7473 & 1.0000 & 0.7343 \\
Cost-weighted Cls.
& 0.7452 & / & 0.7357
& 0.7441 & / & 0.7352 \\
LLMRank 
& 0.7479 & 0.9989 & 0.7346
& 0.7496 & 0.9784 & 0.7361 \\
RouteLLM 
& 0.7475 & 1.0000 & 0.7345
& 0.7473 & 1.0000 & 0.7343 \\
LLM-Blender 
& 0.7489 & 0.9189 & 0.7372
& 0.7500 & 0.9242 & 0.7380 \\
\bottomrule
\end{tabular}
\end{table*}

\begin{table*}[t]
\centering
\small
\setlength{\tabcolsep}{5pt}
\caption{
\textbf{Split sensitivity on RouterEval.}
Comparison under 60/40 and 80/20 train/test splits.
Lower QNC and higher $P_s$ and nAUC are better.
}
\label{tab:split_routereval}
\begin{tabular}{@{}lcccccc@{}}
\toprule
& \multicolumn{3}{c}{60/40 split}
& \multicolumn{3}{c}{80/20 split} \\
\cmidrule(lr){2-4}\cmidrule(l){5-7}
Method
& $P_s \uparrow$ & QNC $\downarrow$ & nAUC $\uparrow$
& $P_s \uparrow$ & QNC $\downarrow$ & nAUC $\uparrow$ \\
\midrule
\textbf{EquiRouter}
& 0.8024 & 0.7987 & 0.6803
& 0.8119 & 0.7903 & 0.6943 \\
kNNRouter
& 0.7844 & 0.8742 & 0.6779
& 0.7929 & 0.8696 & 0.6852 \\
MLPRouter
& 0.7844 & 0.9666 & 0.6693
& 0.7935 & 0.9722 & 0.6756 \\
EmbedLLM
& 0.7992 & 0.8090 & 0.6830
& 0.8083 & 0.8034 & 0.6903 \\
GraphRouter
& 0.7661 & / & 0.6625
& 0.7737 & / & 0.6672 \\
MIRT
& 0.7720 & / & 0.6696
& 0.7586 & / & 0.6661 \\
AvengersPro
& 0.7969 & 0.7767 & 0.6875
& 0.8090 & 0.7826 & 0.6928 \\
CausalRouter
& 0.7969 & 0.9207 & 0.6513
& 0.8029 & 0.9139 & 0.6451 \\
Cost-penalized Reg.
& 0.7837 & 1.0000 & 0.6682
& 0.7841 & / & 0.6749 \\
Cost-weighted Cls.
& 0.3709 & / & 0.3709
& 0.3753 & / & 0.3753 \\
LLMRank 
& 0.7789 & / & 0.6681
& 0.7903 & / & 0.6792 \\
RouteLLM 
& 0.7843 & 1.0000 & 0.6708
& 0.7922 & 1.0000 & 0.6740 \\
\bottomrule
\end{tabular}
\end{table*}

\paragraph{Results.}
On RouterBench, EquiRouter's QNC is 0.9194 under 60/40 and 0.8789 under 80/20.
MIRT and EmbedLLM have lower QNC under 60/40; EquiRouter has the lowest QNC under 80/20.

On MMR-Bench, EquiRouter has the lowest QNC under both splits, decreasing from 0.8733 to 0.8017 as the training fraction increases.
Its peak utility and nAUC remain competitive.

On MixInstruct, QNC is 0.9581 and 0.9361 under the two splits, lower than the other query-based baselines.
LLM-Blender has the lowest QNC among all evaluated methods.
Peak utility and nAUC are comparable across competitive approaches.

On RouterEval, QNC is 0.7987 and 0.7903, and peak utility is 0.8024 and 0.8119.
EquiRouter has the highest peak utility under both splits; AvengersPro has the lowest QNC.

Across all four benchmarks, EquiRouter has lower QNC under 80/20 than under 60/40.
Changes in peak utility and nAUC vary by benchmark.
The two alternative partitions retain competitive quality--cost trade-offs.

\subsection{Sensitivity to the Utility-Indifference Threshold}
\label{app:mixinstruct_epsilon}

MixInstruct uses the original continuous BERTScore utilities, with few exact ties.
CARO introduces cost-aware supervision among similar-utility candidates through threshold $\delta$:

\begin{align}
\mathcal{P}_{\delta}(q)
={}& \{(i,j):u_i(q)-u_j(q)>\delta\}
\nonumber\\
&\cup
\{(i,j):|u_i(q)-u_j(q)|\leq\delta,\,
c_i(q)<c_j(q)\}.
\end{align}

Utility differences above $\delta$ induce higher-utility preferences; within the threshold, lower cost determines preference.
At $\delta=0$, this is exact-equality cost-aware ranking.

\paragraph{Training and threshold selection.}
We use $\delta=0.005$ for MixInstruct and $\delta=0$ for the other three benchmarks.
The MixInstruct objective is

\begin{equation}
\mathcal{L}
=
\mathcal{L}_{\mathrm{MSE}}
+
50\mathcal{L}_{\mathrm{CARO}},
\end{equation}

with unit weights on CARO preference pairs and per-query loss normalization.
Optimization follows the five-epoch, batch-size-2048 Adam schedule with learning rate $10^{-3}$ in Appendix~\ref{app:implementation}.

Training-only validation selects the default threshold from $\{0,0.0025,0.005,0.01,0.02,0.05\}$.
Among configurations whose validation Peak Score is within 0.001 of the maximum, we select the lowest validation QNC.

\paragraph{Sensitivity experiment.}
We evaluate

\[
\delta\in
\{0,0.001,0.0025,0.005,0.0075,0.01,0.02,0.05\}.
\]

on the same 40/60 split (44,000 training and 66,000 test queries) with five optimization seeds (40--44).
Architecture, loss weights, optimizer, training schedule, and cost predictor are fixed.
All thresholds share the best-single-model target $A_{\max}=0.7476113671$, a 100-budget sweep, and the realized-cost interval for nAUC.
Results are means and sample standard deviations across optimization seeds; QNC and saving statistics use runs with defined values.

\begin{table*}[t]
\centering
\small
\setlength{\tabcolsep}{5pt}
\caption{
\textbf{Sensitivity to the utility-indifference
threshold on MixInstruct.}
Results are means $\pm$ standard deviations
over five optimization seeds.
The dagger indicates the default threshold.
Lower QNC and higher $P_s$, saving, and nAUC
are better.
``--'' indicates undefined values.
}
\label{tab:mixinstruct_epsilon}
\resizebox{\textwidth}{!}{%
\begin{tabular}{@{}lcccc@{}}
\toprule
$\delta$
& $P_s$ $\uparrow$
& QNC $\downarrow$
& Saving (\%) $\uparrow$
& nAUC $\uparrow$ \\
\midrule
0
& $0.749869\pm0.000053$
& $0.973393\pm0.005741$
& $2.66\pm0.57$
& $0.736584\pm0.000239$ \\

0.001
& $0.749807\pm0.000031$
& $0.970515\pm0.005845$
& $2.95\pm0.58$
& $0.736639\pm0.000222$ \\

0.0025
& $0.749782\pm0.000072$
& $0.967879\pm0.005150$
& $3.21\pm0.51$
& $0.736710\pm0.000197$ \\

$\mathbf{0.005}^{\dagger}$
& $0.749570\pm0.000147$
& $0.961802\pm0.005831$
& $3.82\pm0.58$
& $0.736776\pm0.000155$ \\

0.0075
& $0.749369\pm0.000244$
& $0.953031\pm0.006654$
& $4.70\pm0.67$
& $0.736880\pm0.000106$ \\

0.01
& $0.749030\pm0.000378$
& $0.947220\pm0.005035$
& $5.28\pm0.50$
& $0.736921\pm0.000099$ \\

0.02
& $0.746588\pm0.000590$
& --
& --
& $0.736449\pm0.000225$ \\

0.05
& $0.714260\pm0.001630$
& --
& --
& $0.713905\pm0.001508$ \\
\bottomrule
\end{tabular}%
}
\end{table*}

\paragraph{Results.}
At $\delta=0.005$, cost-induced near-tie preferences account for 4.042\% of CARO pairs, compared with 0.027\% at $\delta=0$.
Mean QNC decreases from 0.9734 to 0.9618, and Peak Score changes from 0.7499 to 0.7496.
The paired saving increases by $1.16\pm0.27$ percentage points, with a Peak Score difference of $-0.000299\pm0.000150$.

Increasing $\delta$ to 0.01 reduces QNC to 0.9472 and Peak Score to 0.7490.
At $\delta=0.02$ and $\delta=0.05$, QNC is undefined; the latter yields Peak Score 0.7143 and nAUC 0.7139.
Table~\ref{tab:mixinstruct_epsilon} shows the resulting trade-off: small positive thresholds improve cost efficiency with limited utility changes, while larger thresholds sacrifice utility.
The validation-selected $\delta=0.005$ provides the reported quality--cost balance.

\section{Complexity Analysis}
\label{sec:com}

Let \(K\) denote the number of candidate models, \(d_q\) the query
embedding dimension, \(d_m\) the model embedding dimension, and \(D\)
the hidden dimension.
EquiRouter computes a shared query representation
\(z=f(\mathbf{q})\in\mathbb{R}^{D}\) once per query.
For each candidate \(j\), it applies two shared linear projections
\(\phi(m_j)\in\mathbb{R}^{2D}\) and
\(\psi(m_j)\in\mathbb{R}^{D}\), constructs an interaction feature
\(h_j\in\mathbb{R}^{4D}\), and evaluates the shared scoring head
\(g_\theta(h_j)\).

Let \(T_f\) and \(T_g\) denote the computational costs of the query
trunk and the scoring head, respectively.
The routing network has per-query time complexity
\[
O\left(T_f+K(d_mD+D+T_g)\right),
\]
which scales linearly with the number of candidate models.
For a single linear query projection, \(T_f=O(d_qD)\).
Since model embeddings are fixed at inference time, their projections
can also be precomputed, reducing the per-query model-interaction
cost to \(O(K(D+T_g))\).
The complete inference cost additionally includes the query encoder
and the shared cost predictor.

With shared network parameters, model-specific parameter storage is
\(O(Kd_m)\), while batched construction of interaction features requires
\(O(KD)\) temporary memory.
These costs grow linearly with the model pool size and are independent of candidate-model generation, supporting lightweight pre-routing.

\paragraph{Computational Resources.}
Experiments were conducted on NVIDIA A100-SXM4 GPUs (40 GB each), with dual Intel Xeon Platinum 8358 CPUs and approximately 512 GB of RAM. Using cached frozen embeddings, EquiRouter training took approximately 11, 5, 5, and 11 seconds on RouterBench, MMR-Bench, MixInstruct, and RouterEval, respectively, on a single A100 GPU. These measurements include training the utility and cost predictors, including both scoring heads on MMR-Bench. Training across five seeds required approximately 0.044 GPU-hours, excluding feature extraction, initialization, evaluation, and hyperparameter search.

For the additional System One baselines, Clef-Flash was deployed on two A100 GPUs using BF16, with peak allocated GPU memory of approximately 36.7 GiB. Its successful inference runs consumed approximately 83 GPU-hours across four benchmarks, excluding initialization, failed attempts, and exploratory runs. Jev 1.14 was evaluated through an external API.

\section{Limitations and Future Work}
\label{sec:limitations}

\paragraph{Limitations.}
The evaluation uses offline benchmarks with fixed candidate pools and observed utility labels, which may not capture all aspects of live deployment.
Generalization to unseen task groups remains more challenging than adaptation within the evaluated task distributions.

\paragraph{Future work.}
Extending routing to dynamic model pools could incorporate new candidates through lightweight calibration and update cost-aware preferences as prices or availability change.
Uncertainty-aware routing could account for noisy or incomplete utility feedback and distinguish capability advantages from uncertain quality differences.
Extending decision-aligned cost-aware ranking to multi-step agentic workflows could support adaptive model selection across interdependent decisions, optimizing cumulative inference cost while maintaining task-level performance.

\section{Broader Impacts}
\label{sec:impact}

EquiRouter aims to reduce unnecessary inference costs while preserving task utility.
More efficient routing may reduce computational resource consumption and improve access to LLM-based services.

Lower costs may also enable larger-scale deployment, including harmful applications.
Incorrect routing can reduce response quality or reliability when cheaper candidates are insufficient.
Cost efficiency should be evaluated alongside task quality, reliability, and application-specific safety requirements.

The experiments measure inference costs and benchmark utilities; energy consumption and carbon emissions are not directly evaluated.

%% file: checklist.tex
\section*{NeurIPS Paper Checklist}

\begin{enumerate}

\item {\bf Claims}
    \item[] Question: Do the main claims made in the abstract and introduction accurately reflect the paper's contributions and scope?
    \item[] Answer: \answerYes{}
    \item[] Justification: Sections 3--5 and Appendix~\ref{sec:further_analysis} present the analyses and experiments supporting the stated contributions.

\item {\bf Limitations}
    \item[] Question: Does the paper discuss the limitations of the work performed by the authors?
    \item[] Answer: \answerYes{}
    \item[] Justification: Appendix~\ref{sec:limitations} discusses offline evaluation, fixed model pools, and generalization to unseen task groups.

\item {\bf Theory assumptions and proofs}
    \item[] Question: For each theoretical result, does the paper provide the full set of assumptions and a complete (and correct) proof?
    \item[] Answer: \answerNA{}
    \item[] Justification: The paper contains no formal theoretical results.

    \item {\bf Experimental result reproducibility}
    \item[] Question: Does the paper fully disclose all the information needed to reproduce the main experimental results of the paper to the extent that it affects the main claims and/or conclusions of the paper (regardless of whether the code and data are provided or not)?
    \item[] Answer: \answerYes{}
    \item[] Justification: Appendix A documents the experimental protocols, and the released repository provides implementations, configurations, and reproduction scripts.


\item {\bf Open access to data and code}
    \item[] Question: Does the paper provide open access to the data and code, with sufficient instructions to faithfully reproduce the main experimental results, as described in supplemental material?
    \item[] Answer: \answerYes{}
    \item[] Justification: Code and instructions are public at \url{https://github.com/LAMDA-Model-Reuse/EquiRouter}.

\item {\bf Experimental setting/details}
    \item[] Question: Does the paper specify all the training and test details (e.g., data splits, hyperparameters, how they were chosen, type of optimizer) necessary to understand the results?
    \item[] Answer: \answerYes{}
    \item[] Justification: Appendix~\ref{sec:rep} specifies data splits, encoders, cost prediction, optimization settings, and routing protocols.

\item {\bf Experiment statistical significance}
    \item[] Question: Does the paper report error bars suitably and correctly defined or other appropriate information about the statistical significance of the experiments?
    \item[] Answer: \answerYes{}
    \item[] Justification: Table~\ref{tab:multiseed_summary} in Appendix~\ref{app:multiseed} reports means and sample standard deviations over five seeds varying both data splitting and model initialization.

\item {\bf Experiments compute resources}
    \item[] Question: For each experiment, does the paper provide sufficient information on the computer resources (type of compute workers, memory, time of execution) needed to reproduce the experiments?
    \item[] Answer: \answerYes{}
    \item[] Justification: Appendix~\ref{sec:com} reports hardware specifications, EquiRouter training times, and System One inference resource measurements.
    
\item {\bf Code of ethics}
    \item[] Question: Does the research conducted in the paper conform, in every respect, with the NeurIPS Code of Ethics \url{https://neurips.cc/public/EthicsGuidelines}?
    \item[] Answer: \answerYes{}
    \item[] Justification: The study uses existing benchmarks and cached model outcomes without collecting new participant data.

\item {\bf Broader impacts}
    \item[] Question: Does the paper discuss both potential positive societal impacts and negative societal impacts of the work performed?
    \item[] Answer: \answerYes{}
    \item[] Justification: Appendix~\ref{sec:impact} discusses accessibility and resource benefits, misuse risks, and consequences of incorrect routing.
    
\item {\bf Safeguards}
    \item[] Question: Does the paper describe safeguards that have been put in place for responsible release of data or models that have a high risk for misuse (e.g., pre-trained language models, image generators, or scraped datasets)?
    \item[] Answer: \answerNA{}
    \item[] Justification: No high-risk datasets or generative model weights are released.

\item {\bf Licenses for existing assets}
    \item[] Question: Are the creators or original owners of assets (e.g., code, data, models), used in the paper, properly credited and are the license and terms of use explicitly mentioned and properly respected?
    \item[] Answer: \answerYes{}
    \item[] Justification: The original resources are cited, and third-party licenses and usage terms are documented in the
released repository.

\item {\bf New assets}
    \item[] Question: Are new assets introduced in the paper well documented and is the documentation provided alongside the assets?
    \item[] Answer: \answerYes{}
    \item[] Justification: The released EquiRouter code includes installation, training, evaluation, citation, and license documentation.

\item {\bf Crowdsourcing and research with human subjects}
    \item[] Question: For crowdsourcing experiments and research with human subjects, does the paper include the full text of instructions given to participants and screenshots, if applicable, as well as details about compensation (if any)? 
    \item[] Answer: \answerNA{}
    \item[] Justification: The study includes no crowdsourcing or human-subject experiments.

\item {\bf Institutional review board (IRB) approvals or equivalent for research with human subjects}
    \item[] Question: Does the paper describe potential risks incurred by study participants, whether such risks were disclosed to the subjects, and whether Institutional Review Board (IRB) approvals (or an equivalent approval/review based on the requirements of your country or institution) were obtained?
    \item[] Answer: \answerNA{}
    \item[] Justification: The study involves no human participants.

\item {\bf Declaration of LLM usage}
    \item[] Question: Does the paper describe the usage of LLMs if it is an important, original, or non-standard component of the core methods in this research? Note that if the LLM is used only for writing, editing, or formatting purposes and does \emph{not} impact the core methodology, scientific rigor, or originality of the research, declaration is not required.
    \item[] Answer: \answerYes{}
    \item[] Justification: Appendix~\ref{app:implementation} describes Qwen3-8B model-description generation, and Appendix~\ref{app:system_one_protocol} specifies Jev 1.14 and Clef-Flash routing.

\end{enumerate}